\def\arxivversion{}
\documentclass{article}
\usepackage{iclr2027_conference,times}
\ifdefined\arxivversion\iclrfinalcopy\fi

\usepackage[utf8]{inputenc}
\usepackage[T1]{fontenc}
\usepackage{amsmath,amssymb,amsthm}
\usepackage{amsfonts}
\usepackage{stmaryrd}   \usepackage{mathtools}
\usepackage{bbm}        \usepackage{graphicx}
\graphicspath{{images/}}
\usepackage{hyperref}
\usepackage{xcolor}
\usepackage{tikz}
\usetikzlibrary{calc,patterns,positioning,arrows.meta,fit,backgrounds}
\usepackage{pgfplots}
\pgfplotsset{compat=1.18}
 \usepackage{todonotes}
\definecolor{turquoise}{RGB}{64,224,208}
\usepackage{float}
\usepackage{microtype}
\usepackage{array}
\usepackage{booktabs}
 \usepackage{enumitem}

\newtheorem{definition}{Definition}[section]

\newcommand{\A}{\mathcal{A}}
\newcommand{\F}{\mathcal{F}}

\newcommand{\Z}{\mathbb{Z}}
\newcommand{\llb}{\llbracket}
\newcommand{\rrb}{\rrbracket}

\definecolor{cRegion}{RGB}{78,121,167}    \definecolor{cAdj}{RGB}{225,87,89}        \definecolor{cAdjCell}{RGB}{76,76,76}     \definecolor{cParity}{RGB}{89,161,79}     \definecolor{cShift}{RGB}{242,142,43}     \definecolor{cAlt1}{RGB}{148,103,189}     \definecolor{cAlt2}{RGB}{60,150,155}      \definecolor{cAlt3}{RGB}{140,86,75}       \newcommand{\noinfo}{no-information}
\newcommand{\descsetup}{description-based}
\newcommand{\exsetup}{example-based}
\newcommand{\splitmixed}{mixed}
\newcommand{\splitaxis}{axis}
\newcommand{\splitrowpair}{row-pair}

\title{Knowledge transfer between simple discrete worlds}

\ifdefined\arxivversion
\makeatletter
\def\@maketitle{\vbox{\hsize\textwidth
{\LARGE\sc \@title\par}
\vskip 0.3in
{\centering\def\And{\end{tabular}\hspace{4em}\begin{tabular}[t]{c}}\begin{tabular}[t]{c}\@author\end{tabular}\par}
\vskip 0.3in minus 0.1in}}
\makeatother
\author{\textbf{Silv\`ere Gangloff}\\
University of Ostrava\\
\texttt{silvere.gangloff@osu.cz}
\And \textbf{Jan H\r{u}la}\\
University of Ostrava\\
\texttt{jan.hula@osu.cz}}
\else
\author{Silv\`ere Gangloff \And Jan H\r{u}la}
\fi

\begin{document}
\maketitle
\ifdefined\arxivversion\lhead{Preprint}\fi

\begin{abstract}
Adapting to a new world and reasoning correctly about which outcomes are possible under rules one has not met before is a basic capacity underlying general intelligence. We study it in small discrete worlds: sets of grid configurations consistent with a shared set of rules, concretely Sudoku-like puzzles built from a common base and a handful of extra rules. We train a single depth-recurrent model on many combinations of rules and test it at four levels of novelty: a new puzzle, a new combination of known rules, a pairing of rules never seen together, and a rule instance not seen at all during training. The model receives the active rules either as a symbolic description or as two solved examples, and a model that gets no rule information is used as a baseline. At recombining rules it was trained on, the description matches or exceeds examples ($99\%$ against $95\%$ on new combinations). On rule instances whose components never co-occurred in training, the description falls below the no-information baseline while examples stay above it. The framework allows us to test many such generalization regimes in more systematic way than benchmarks such as ARC-AGI and provides new ground for evaluating depth-recurrent reasoning models, which are typically tested on saturated Sudoku benchmarks.
\end{abstract}
 
\section{Introduction}
\label{sec:introduction}

A recurring puzzle in artificial intelligence is the mismatch between domains of competence: systems that exceed human performance on tasks perceived as difficult often fail on tasks that appear far simpler. Following \citet{chollet2019arc}, this points to a distinction between \emph{narrow intelligence} (high performance on a specific, well-represented task distribution) and \emph{general intelligence} (the capacity to adapt to, and learn the structure of, previously unseen environments). ARC (the Abstraction and Reasoning Corpus,~\citealp{chollet2019arc}) operationalizes this via small sets of input--output examples from which a hidden transformation must be inferred rather than memorized, but its tasks remain shaped by human visual conventions (connected-component segmentation, color-as-category symbolism), so strong performance may partly reflect familiarity with these conventions, which makes such studies hard to control.

We adopt a more minimal formalization, close to the intuitive notion of a \emph{world}: a finite grid governed by a small set of explicit rules, each forbidding certain patterns (Section~\ref{sec:small-worlds} makes this formal). Physical laws can be phrased the same way, and familiar puzzles such as Sudoku and mazes are special cases with a particular rule vocabulary. Understanding a world then means compressing the dependencies between its patterns into a few rules that are simple to store and apply, the core capability of a \emph{world model}~\citep{ha2018worldmodels,lecun2022path}. Because these rules come from the structure of the world itself rather than from a human-designed visual vocabulary, the resulting benchmark is less human-centric than ARC; this matters because general intelligence need not be human-like, and could exceed human intelligence in ways a human-anchored benchmark would fail to capture.

We study how knowledge \emph{transfers} across different worlds, for a relatively small \emph{parameterized} family of worlds: given a model trained on many combinations of rules from a fixed pool, does it correctly apply an unseen \emph{combination} of already-known rules and, more strictly, a rule \emph{instance} it has never encountered? We study this in a family of Sudoku-like grid worlds: a fixed Latin-square base plus a few extra rules drawn from a fixed pool (Section~\ref{sec:experimental-setup}). To do so, we train a depth-recurrent model (similar to those recently used on Sudoku~\citep{wang2025hrm}) on a puzzle together with a context that carries the information about the active rules: either two solved examples from which the rules can be inferred, or the rules stated explicitly as a short symbolic description. A model given no such context serves as the no-information baseline. 

We test four levels of novelty, from a new puzzle in a trained world to a rule instance never seen at all. The two ways of communicating the rules turn out equally good at recombining rules the model was trained on. On rule instances whose components were never seen together in training, the symbolic description falls below the no-information baseline while examples stay above it. Section~\ref{sec:results} shows how this relates to how much co-occurring rule structure the training set exposes. 

The generators from the proposed setup are easily extensible to more complex rule structures and worlds and can therefore replace already saturated fixed-game benchmarks for recurrent reasoning models~\citep{jolicoeurmartineau2025trm,davis2026latticededuction} such as Sudoku-Extreme~\citep{wang2025hrm}.

\paragraph{Contributions.} \ifdefined\arxivversion Code and data are available at \url{https://github.com/Sfgangloff/GridWorldHyperNetworks}.\else Code and data are available at an anonymized repository.\footnote{\url{https://anonymous.4open.science/r/GridWorldHyperNetworks-E993/}}\fi This paper provides:
\begin{enumerate}[itemsep=1pt,topsep=2pt,parsep=1pt,partopsep=0pt,leftmargin=*]
  \item A formalization of a setup in which several distinct modes of novelty can be studied (Section~\ref{sec:worlds-framework}).
  \item A parameterized family of Sudoku-like grid worlds whose rules decompose into reusable components (Section~\ref{sec:family}).
  \item An empirical characterization of where compositional generalization breaks down and how it depends on how rule identity is communicated (Section~\ref{sec:results}).
  \item Evidence that the limiting factor is exposure to co-occurring rule structure rather than model capacity or the symbolic format (Section~\ref{sec:results}, Appendix~\ref{app:results}).
\end{enumerate}
 
\section{Parameterized families of small discrete worlds}
\label{sec:worlds-framework}

\newcommand{\gridlines}{\foreach \x in {0,...,6} \draw[gray!50,very thin] (\x,0) -- (\x,6);
  \foreach \y in {0,...,6} \draw[gray!50,very thin] (0,\y) -- (6,\y);
  \draw[thick] (0,0) rectangle (6,6);}
\newcommand{\boxlines}{\draw[thick] (3,0) -- (3,6); \draw[thick] (0,2) -- (6,2); \draw[thick] (0,4) -- (6,4);}
\newcommand{\gridvalues}[1]{\foreach \v [count=\i from 0] in {#1} {\pgfmathtruncatemacro{\r}{int(\i/6)}\pgfmathtruncatemacro{\c}{mod(\i,6)}\ifnum\v>0 \node[font=\scriptsize] at (\c+0.5,5.5-\r) {\v};\fi}}
\newcommand{\ruleglyph}[1]{\raisebox{2.5pt}{\begin{tikzpicture}[scale=0.19,baseline={(current bounding box.center)}]
    \useasboundingbox (0,0) rectangle (6,6);
    #1\gridlines
  \end{tikzpicture}}}
 
This section sets up the formal framework: worlds as sets of configurations avoiding forbidden patterns (Section~\ref{sec:small-worlds}), and parameterized families of worlds indexed by combinations of rules from a fixed pool (Section~\ref{sec:parameterized-families}), instantiated concretely in Section~\ref{sec:experimental-setup}.

\subsection{Small discrete worlds}
\label{sec:small-worlds}

Let $\A$ be a finite alphabet and $N \ge 1$ an integer. We call \emph{configuration} any element of $\A^{\llb 1,N\rrb^2}$, i.e.\ any coloring of the $N \times N$ grid with symbols from $\A$. A \emph{pattern} is a function $p : \mathbb{U} \to \A$, where $\mathbb{U} \subset \Z^2$ is a finite set called the \emph{support} of $p$. A pattern $p$ appears in a configuration $x \in \A^{\llb 1,N\rrb^2}$ if some shifted copy of $p$ is found in $x$: there is $v \in \Z^2$ such that $x_{u+v} = p_u$ for every $u \in \mathbb U$.

\begin{definition}[World]
\label{def:world}
Let $\F$ be a finite set of patterns on alphabet $\A$. The \emph{world} with parameters $(\A, N, \F)$ is $W_N(\A,\F) = \{ x \in \A^{\llb 1,N\rrb^2} \mid \text{no pattern in } \F \text{ appears in } x \}$. Elements of $W_N(\A,\F)$ are called \emph{admissible configurations}, and $\F$ the world's \emph{forbidden patterns}.
\end{definition}

This forbidden-pattern formalization is the finite-grid analogue of the subshift of finite type of symbolic dynamics~\citep{lind1995introduction}. We impose no restriction on pattern support size relative to $N$: several of our family's rules (Section~\ref{sec:experimental-setup}) forbid patterns whose support spans an entire row, column, or box.

\subsection{Parameterized families}
\label{sec:parameterized-families}

To study generalization, we consider \emph{families} of worlds sharing the same alphabet and grid size, differing only in which rules are active, and study generalization through different train--test splits over such a family.

\begin{definition}[Parameterized family]
\label{def:parameterized-family}
Let $\mathcal{P} = \{\F_1, \ldots, \F_K\}$ be a finite \emph{pool} of forbidden sets, each $\F_k$ a finite set of patterns defining one \emph{rule}. A \emph{combination} is a subset $S \subseteq \{1,\ldots,K\}$, giving rise to the world $W_N\bigl(\A,\, \bigcup_{k \in S} \F_k\bigr)$. A \emph{parameterized family} associated to $\mathcal{P}$ is any collection $\{\, W_N(\A,\, \textstyle\bigcup_{k \in S} \F_k) \mid S \in \mathcal{S} \,\}$, $\mathcal{S} \subseteq 2^{\{1,\ldots,K\}}$, of worlds indexed by some chosen subset $\mathcal{S}$ of all possible combinations of rules from $\mathcal{P}$.
\end{definition}

This shared structure is what makes it meaningful to ask whether a model trained on some worlds generalizes to a new one - a new combination $S \in \mathcal{S}$ of already-known rules, or, more strictly, to a world with a genuinely new rule outside $\mathcal{P}$ altogether.

\subsection{Completion problems}
\label{sec:puzzles-def}

Deciding whether a full configuration is admissible is a constraint-satisfaction verification, and every world in our family is a CSP in this sense (Section~\ref{sec:related-work}). Beyond that verification, a natural task in a world $W_N(\A,\F)$ is to recover a full configuration from a partial one - itself a CSP, of finding a satisfying assignment rather than merely verifying one.

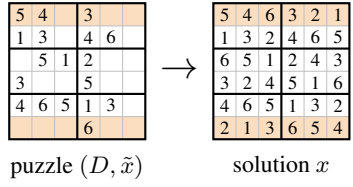
\begin{figure}[t]
\centering
\begin{tikzpicture}[font=\footnotesize]
  \newcommand{\linkshade}{\fill[cShift!30] (0,5) rectangle (6,6); \fill[cShift!30] (0,0) rectangle (6,1);}
  \begin{scope}[scale=0.3]
    \linkshade\gridlines\boxlines
    \gridvalues{5,4,0,3,0,0, 1,3,0,4,6,0, 0,5,1,2,0,0, 3,0,0,5,0,0, 4,6,5,1,3,0, 0,0,0,6,0,0}
    \node[anchor=north] at (3,-0.3) {puzzle $(D,\tilde x)$};
  \end{scope}
  \node[font=\Large] at (2.25,0.9) {$\rightarrow$};
  \begin{scope}[scale=0.3, shift={(9,0)}]
    \linkshade\gridlines\boxlines
    \gridvalues{5,4,6,3,2,1, 1,3,2,4,6,5, 6,5,1,2,4,3, 3,2,4,5,1,6, 4,6,5,1,3,2, 2,1,3,6,5,4}
    \node[anchor=north] at (3,-0.3) {solution $x$};
  \end{scope}
\end{tikzpicture}
\caption{A completion problem (Definition~\ref{def:puzzle}): given cells $D$ (left) determine a unique admissible configuration $x$ (right) in some world $W_N(\A,\F)$. This particular puzzle comes from the concrete family introduced in Section~\ref{sec:experimental-setup}.}
\label{fig:completion-problem}
\end{figure}

\begin{definition}[Completion problem]
\label{def:puzzle}
Let $W_N(\A,\F)$ be a world. A \emph{partial configuration} is a pair $(D, \tilde x)$ with $D \subseteq \llb 1,N\rrb^2$ and $\tilde x : D \to \A$; cells in $D$ are \emph{given}, cells outside $D$ are \emph{blank}. A partial configuration $(D,\tilde x)$ is a \emph{completion problem} for $W_N(\A,\F)$ if there is exactly one admissible configuration $x \in W_N(\A,\F)$ with $x|_D = \tilde x$; this unique $x$ is the problem's \emph{solution}, and \emph{solving} it means recovering $x$ from $(D,\tilde x)$ and the world's rules $\F$. Figure~\ref{fig:completion-problem} shows one concrete instance.
\end{definition}

Sudoku, and grid puzzles generally, are exactly completion problems for a world whose forbidden patterns encode the puzzle's rules; the fewer cells given, the more the rules alone must determine the rest of the grid, so the size of $D$ is a coarse proxy for difficulty. 
\section{Experimental setup}
\label{sec:experimental-setup}

We instantiate the framework of Section~\ref{sec:worlds-framework} with a family of Sudoku-like worlds (Section~\ref{sec:family}), three ways of telling the model which rules are active (Section~\ref{sec:how-rules-communicated}), four levels of novelty at test time (Section~\ref{sec:testing-different-modes-of-novelty}) and a dial that controls how much compositional structure the training set exposes (Section~\ref{sec:exposure}). Figure~\ref{fig:overview} summarizes this setup.

\subsection{Sudoku-style parameterized family of worlds}
\label{sec:family}

Every world in our family is subject to a fixed set $\mathcal{L}$ of \emph{base} rules, the Latin-square constraint that every row and every column contains each symbol exactly once, together with $2$--$4$ additional rules drawn from a \emph{variable} pool $\mathcal{V}$ of $49$ rules; in the terms of Definition~\ref{def:parameterized-family}, $\mathcal{P} = \mathcal{L} \cup \mathcal{V}$ and every combination $S$ contains all of $\mathcal{L}$. We use $6\times6$ grids with symbols $\{1,\ldots,6\}$, the smallest size divisible by both $2$ and $3$ as several pool rules tile the grid in $2\times3$ or $3\times2$ blocks, which keeps a sweep over thousands of rule combinations tractable without making puzzles trivial (the \noinfo{} baseline of Section~\ref{sec:results} reaches only $61\%$ in-distribution accuracy on our main split). Nothing in the formalization is specific to this size.

Holding the base fixed serves two purposes. Any generalization gap then reflects the variable pool alone, not a second source of novelty; and because the base constrains every cell through its row and column while the variable rules are mostly local (one box, one diagonal, a pair of rows), many cells can be blanked while a single valid completion remains, and the solver used to sample puzzles can check uniqueness quickly (Section~\ref{sec:puzzles}). The pool rules are chosen for their similarity to the base and to variants familiar from Sudoku (boxes, diagonals, windows), each still introducing some difference, so the pool stays one recognizable family rather than an arbitrary set. Most of the rules in the pool decompose into shared, reusable components (template type, axis, position, parameter), and this lets us hold out a rule while keeping its components in the training set. Table~\ref{tab:pool6-families} lists the rule types. Appendix~\ref{app:design-choices} gives the rationale for these design choices.

\begin{table}[t]
\centering
\caption{Rule types in the $49$-instance pool $\mathcal{V}$. Every rule is a set of forbidden patterns on the $6\times6$ grid; the glyph shades the cells one instance of the type constrains. Full per-instance definitions in Appendix~\ref{app:pool6}.}
\label{tab:pool6-families}
\footnotesize
\begin{tabular}{@{}>{\centering\arraybackslash}m{1.3cm} >{\raggedright\arraybackslash}m{3.6cm} >{\centering\arraybackslash}m{0.6cm} >{\raggedright\arraybackslash}m{0.42\linewidth}@{}}
\toprule
 & Type & \# & Description of the rules of this type in the family \\
\midrule
\ruleglyph{\fill[cRegion!35] (0,4) rectangle (3,6); \boxlines} & Unique-occurrence regions - \footnotesize\itshape a region where each symbol appears at most once & 11 &
  A fixed support (a $2\times3$ box, its scattered counterpart, a diagonal, a window, the four center or corner cells) contains each symbol at most once. \\
\addlinespace
\ruleglyph{\fill[cAdjCell!60] (2,3) rectangle (3,4); \foreach \p in {(1,5),(3,5),(0,4),(4,4),(0,2),(4,2),(1,1),(3,1)} {\fill[cAdj!35] \p rectangle ++(1,1);}} & Adjacency - \footnotesize\itshape cells in a fixed spatial relation must differ & 2 &
  Two cells a knight's move apart (\texttt{AntiKnight}) or diagonally adjacent (\texttt{AntiKing}) never share a value. \\
\addlinespace
\ruleglyph{\foreach \i in {0,...,5} {\fill[cParity!35] (\i,5-\i) rectangle ++(1,1);} \node[font=\tiny] at (4.5,4.5) {odd};} & Value-restricted regions - \footnotesize\itshape a region confined to a fixed subset of values & 4 &
  Every cell of a diagonal takes values in a fixed set $V \subseteq \{1,\ldots,6\}$ (odd or even). \\
\addlinespace
\ruleglyph{\fill[cShift!35] (0,5) rectangle (6,6); \fill[cShift!35] (0,0) rectangle (6,1); \draw[-{Stealth[length=3pt]},cShift!80!black,thick] (3,4.8) -- (3,1.2); \node[font=\tiny,fill=white,inner sep=0.5pt] at (3,3) {$+k$};} & Value-shift relations - \footnotesize\itshape two regions whose values are related by a fixed offset & 32 &
  A pair of rows or columns $a,b$ (one specific pair, or all three pairs of the top and bottom halves) satisfies $b = a + k \bmod 6$ cell by cell, $k \in \{1,\ldots,5\}$. \\
\bottomrule
\end{tabular}
\end{table}
 
\paragraph{Puzzles.}
\label{sec:puzzles}
Not every combination of rules admits a configuration, and a world with none cannot be turned into a puzzle. A combination is retained only if a custom solver (Appendix~\ref{app:generation}) samples enough full configurations from it that stay uniquely completable with at least $14$ of the $36$ cells blanked, about the fraction blanked in a comparably hard $9\times9$ Sudoku. To obtain one puzzle we sample an admissible configuration, then blank cells one at a time, rejecting any removal that makes the remainder ambiguous, until a calibrated target blank count is reached. We generate $500$ training worlds with $2000$ training and $200$ held-out puzzles each. The held-out puzzles of $50$ of these worlds form the validation set used to select checkpoints, and also serve as the in-distribution test set (level L0 of Section~\ref{sec:testing-different-modes-of-novelty}). The other three levels are evaluated on $30$ further worlds each, using their $200$ held-out puzzles. Figure~\ref{fig:completion-problem} shows one such puzzle and its solution.

\subsection{How the rules are communicated}
\label{sec:how-rules-communicated}

A partial configuration alone need not determine its solution when the rules are unknown, so the model must be told, or must infer, which rules are active. We compare three setups (Table~\ref{tab:setups}), differing only in the prefix placed before the $36$ puzzle cells.

\begin{table}[t]
\centering
\caption{The three conditioning setups. Only the prefix placed before the $36$ puzzle cells differs.}
\label{tab:setups}
\footnotesize
\begin{tabular}{@{}l l r r >{\raggedright\arraybackslash}p{0.36\linewidth}@{}}
\toprule
Setup & Prefix & Len. & Symbols used & What the model must do \\
\midrule
\noinfo{}     & none                           & 36  & 7  & Infer the rules from the given cells alone; lower bound on how much rule information the puzzle itself carries. \\
\addlinespace
\descsetup{}  & $4$ rules $\times$ $4$ tokens  & 52  & 36 & Read the rule identity as symbols and apply it. \\
\addlinespace
\exsetup{}    & $2$ solved grids ($72$ tokens) & 108 & 7  & Infer the rules from the examples, then apply them; mirrors the ARC-AGI format. \\
\bottomrule
\end{tabular}
\end{table}
 
In the \descsetup{} setup, each active rule is represented using four tokens, one for each of the following slots: template type, axis, position, and parameter. If a given rule does not use some slots, or if fewer than four variable rules are present (as worlds have at most four variable rules), the unused slots are filled with a special null token. In this setup, the puzzle prefix has therefore a constant length of 16 tokens. The base rules are never explicitly described. For the world of Figure~\ref{fig:completion-problem}, $\{\texttt{BoxesRect2x3},\ \texttt{RowLink(1,6,}k{=}3)\}$, this description reads
\[
\langle\texttt{box},\ \varnothing,\ \texttt{2x3},\ \varnothing\rangle\ \
 \langle\texttt{link},\ \texttt{row},\ \texttt{(1,6)},\ \texttt{3}\rangle\ \
 \langle\varnothing,\varnothing,\varnothing,\varnothing\rangle\ \
 \langle\varnothing,\varnothing,\varnothing,\varnothing\rangle
\]
(slot names shown here for clarity are schematic; $\varnothing$ is the null token).

The description alphabet adds $29$ symbols to the $7$ puzzle symbols. In the \exsetup{} setup the context is two solved puzzles from the same world. Because the token-mixing layers grow with sequence length, we also report a control with the description padded to the \exsetup{} length (Appendix~\ref{app:padding}). Rules fill the four slots in alphabetical order of their identifier (Appendix~\ref{app:pool6}).

\subsection{Four levels of novelty}
\label{sec:testing-different-modes-of-novelty}

\begin{figure}[t]
\centering
\begin{tikzpicture}[font=\footnotesize, every node/.style={inner sep=1pt}]
\node[anchor=south west,font=\footnotesize\itshape] at (0,1.95) {Rules list};
\node[anchor=west,align=left,text width=3.1cm] at (0,0.84) {Latin square (base)\\
  $+$ \textcolor{cRegion!80!black}{\texttt{BoxesRect2x3}}\\
  $+$ \textcolor{cParity!80!black}{\texttt{MainDiagonal}}\\
  $+$ \textcolor{cShift!80!black}{\texttt{RowLink(1,6,$k$)}}};
\node[anchor=south west,font=\footnotesize\itshape] at (3.4,1.95) {Rules schema};
\begin{scope}[xshift=3.4cm,scale=0.28]
  \fill[cRegion,opacity=0.35] (0,4) rectangle (3,6);
  \fill[cParity,opacity=0.35] (0,5) rectangle (1,6); \fill[cParity,opacity=0.35] (1,4) rectangle (2,5); \fill[cParity,opacity=0.35] (2,3) rectangle (3,4);
  \fill[cParity,opacity=0.35] (3,2) rectangle (4,3); \fill[cParity,opacity=0.35] (4,1) rectangle (5,2); \fill[cParity,opacity=0.35] (5,0) rectangle (6,1);
  \fill[cShift,opacity=0.35] (0,5) rectangle (6,6); \fill[cShift,opacity=0.35] (0,0) rectangle (6,1);
  \gridlines\boxlines
  \draw[-{Stealth[length=4pt]},cShift!80!black,thick] (6.6,5.5) -- (6.6,0.5) node[midway,right,font=\scriptsize,text=black] {$+k$};
\end{scope}
\node[anchor=north,font=\footnotesize\bfseries] at (2.8,-0.3) {(a) example world};

\begin{scope}[shift={(6.4,0.7)}, font=\scriptsize]
\newcommand{\strip}[5]{\draw[fill=#4] (#1,#2) rectangle (#1+#3,#2+0.3); \node at (#1+#3/2,#2+0.15) {#5};}
  \node[anchor=west,font=\footnotesize] at (-0.1,1.6) {\noinfo};
  \strip{0}{1.1}{1.8}{black!10}{puzzle}
  \node[anchor=west,font=\footnotesize] at (-0.1,0.7) {\descsetup};
  \strip{0}{0.2}{1.7}{cRegion!30}{description}
  \strip{1.7}{0.2}{1.8}{black!10}{puzzle}
  \node[anchor=west,font=\footnotesize] at (-0.1,-0.2) {\exsetup};
  \strip{0}{-0.7}{1.8}{cShift!25}{example} \strip{1.8}{-0.7}{1.8}{cShift!25}{example} \strip{3.6}{-0.7}{1.8}{black!10}{puzzle}
  \node[anchor=north,font=\footnotesize\bfseries] at (2.7,-1.0) {(b) how the active rules are communicated};
\end{scope}

\begin{scope}[shift={(0.05,-1.4)}, every node/.style={inner sep=1pt}, font=\scriptsize]
\fill[gray!12] (-0.9,-0.25) rectangle (12.6,0.15);
  \fill[gray!12] (-0.9,-1.05) rectangle (0,0.15);
\foreach \x in {-0.9,0,3.2,6.8,9.3,12.6} \draw[gray!50,thin] (\x,0.15) -- (\x,-1.05);
  \foreach \y in {0.15,-0.25,-0.65,-1.05} \draw[gray!50,thin] (-0.9,\y) -- (12.6,\y);
\node[anchor=west,font=\bfseries\footnotesize] at (0.1,-0.05) {L0 (familiar world)};
  \node[anchor=west,font=\bfseries\footnotesize] at (3.3,-0.05) {L1 (new combination)};
  \node[anchor=west,font=\bfseries\footnotesize] at (6.9,-0.05) {L2 (new pairing)};
  \node[anchor=west,font=\bfseries\footnotesize] at (9.4,-0.05) {L3 (new rule instance)};
\node[anchor=west,font=\itshape] at (-0.8,-0.45) {train};
  \node[anchor=west,font=\itshape] at (-0.8,-0.85) {test};
\node[anchor=west] at (0.1,-0.45) {$\{A,B,C\}$, $\{A,D\}$};
  \node[anchor=west] at (3.3,-0.45) {$\{A,B,C\}$, $\{A,D\}$, $\{B,D\}$};
  \node[anchor=west] at (6.9,-0.45) {$(A,B)$, $(C,D)$};
  \node[anchor=west] at (9.4,-0.45) {$(A,B)$, $(B,C)$, $(C,D)$};
\node[anchor=west] at (0.1,-0.85) {$\{A,B,C\}$, new puzzle};
  \node[anchor=west] at (3.3,-0.85) {$\{A,B,D\}$};
  \node[anchor=west] at (6.9,-0.85) {$(A,C)$};
  \node[anchor=west] at (9.4,-0.85) {$(A,H)$};
  
  \node[anchor=north,font=\footnotesize\bfseries] at (6.7,-1.2) {(c) forms of novelty};
\end{scope}
\end{tikzpicture}
\caption{Overview of the setup. (a) Example world: list of rules and schematic representation of these rules. (b) How the rules are communicated: information is appended before the puzzle under diverse forms. (c) Illustration of the forms of novelty. Letters designate rules.}
\label{fig:overview}
\end{figure}
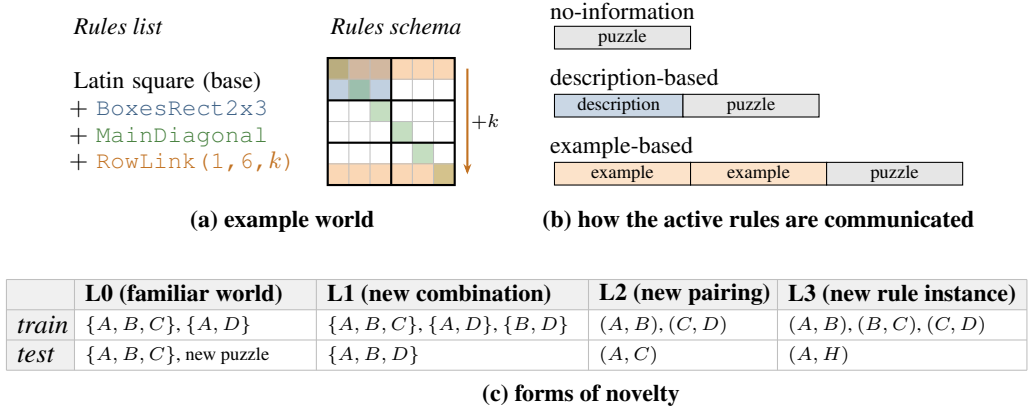

A model may generalize to a new puzzle in several qualitatively different ways: the puzzle alone may be new, or the world itself, through a new combination of trained rules, a pairing of rules never seen together, or a rule instance never seen at all (Figure~\ref{fig:overview}c). The parameterized family lets us separate these cases instead of conflating them in one held-out set.

\paragraph{New configuration in a familiar world (L0).}
The most basic form of novelty is when considering an unseen configuration from a world that was already present during training. This is the standard form of generalization tested by puzzle benchmarks. We report this as our in-distribution reference.

\paragraph{New combination of familiar rules (L1).}
We next test whether the model can solve puzzles from a world whose complete combination of variable rules was never used during training, while every individual rule in that combination was present in the training set. We refer to this as \emph{combination generalization}.

\paragraph{New co-occurrence of familiar rules (L2).}
A new rule combination does not necessarily imply that the interactions within it are new.
Even if the complete set of active rules has never appeared during training, each of its
pairs may have appeared in some other training world. We therefore consider a stronger
setting in which whole sets of rule pairs are deliberately prevented from co-occurring
during training. We partition the rule pool into two groups and remove from training every
world that contains rules from both of them: each rule is still individually represented and
pairs within a group still occur, but the model never observes a world in which a
cross-group pair is active together. At test time it must solve worlds that contain exactly
such previously unseen co-occurrences.

This construction violates the \emph{connected support} condition of \citet{schug2024modular}, under which any two rules are linked by a chain of training worlds in which consecutive rules co-occur; in this split, which we call \emph{no-mix}, the support falls into two disconnected components. Section~\ref{sec:exposure} turns the degree of connectedness into a controlled variable.

\paragraph{New rule instance (L3).}
Finally, we test generalization to worlds containing a rule instance that was never present during training. The rule pool is parameterized so that many rule instances can be decomposed into components shared with other rules. We can therefore hold out a complete rule while retaining its constituent components elsewhere in the training set. For example, the model sees value-shift rules linking rows $1$ and $6$ with offsets $k=1$ and $k=2$, and offset $k=3$ on other row pairs, and is tested on the rule linking rows $1$ and $6$ with $k=3$. Solving puzzles in such worlds requires applying the structure of a rule type to an instance never encountered as a whole, not merely recombining known rules. Whether this is possible should depend on which of the instance's components were seen together: we call a held-out instance \emph{supported} if every pair of its components (for example, the link type together with the column axis) occurs in some trained rule, and \emph{unsupported} if at least one such pair never occurs in training. Section~\ref{sec:exposure} varies the support of a fixed set of held-out instances directly; Appendix~\ref{app:results-splits} repeats the L3 evaluation on three further held-out sets with three draws each.

\subsection{Controlling exposure to compositional structure}
\label{sec:exposure}
The levels above fix \emph{what} is new at test time. We additionally control \emph{how much} relevant compositional structure the training set exposes, with one dial for L2 and one for L3. The L2 dial starts from the no-mix split and reintroduces training worlds that mix the two rule groups at a graded rate, from none up to the rate at which they occur naturally. The L3 dial holds out a fixed set of column links and varies how many training worlds contain a supporting instance (another column link), from none, where the held-out instances are unsupported, up to the natural rate. In both, the test worlds are identical at every level and the training sets are nested, so any change in accuracy reflects exposure alone, and a \noinfo{} model is trained at every level so that each number has its floor beside it. Level values, pair coverages and the full curves are given in Appendix~\ref{app:exposure}.

\subsection{Model and training}
\label{sec:model}
\label{sec:architecture}
All setups share one depth-recurrent model that reads the token sequence and predicts the completed grid. It follows the Hierarchical Reasoning Model recipe~\citep{wang2025hrm}: two coupled latent states updated at different rates, a one-step gradient approximation in place of backpropagation through the full recursion, deep supervision, and a learned halting signal. The block differs: instead of self-attention we use an MLP-Mixer-style block~\citep{tolstikhin2021mlpmixer}, one SwiGLU MLP mixing over sequence positions and one over channels, each with a residual connection and post-RMSNorm. Because the token-mixing weights are tied to absolute positions, no positional encoding is needed. The model has hidden size $256$ and $2$ blocks, with the same architecture and hyperparameters in every setup; only the input sequence differs, and with it the size of the token-mixing MLP, which gives $1.09$M, $1.11$M and $1.25$M parameters for the \noinfo{}, \descsetup{} and \exsetup{} setups respectively (Appendix~\ref{app:padding} controls for this difference). We train every setup and split with the same AdamW recipe for $20$ epochs of $500$ batches of $512$ puzzles ($10{,}000$ batches with $8$ optimizer steps each, about five passes over the training puzzles), with a $200$-step warmup followed by a constant learning rate of $10^{-4}$, keep the checkpoint with the best validation exact-match accuracy, and report exact-match accuracy, the fraction of puzzles whose $36$ cells are all correct, as mean $\pm$ standard deviation over $4$ training seeds for every run. Where a result is replicated on several draws of a split we report each draw separately, and for the headline comparison we give a paired bootstrap confidence interval over test worlds (Appendix~\ref{app:results-splits}).
Appendix~\ref{app:model} gives the full specification and training details. Figure~\ref{fig:overview} summarizes this full setup.
 
\section{Results}
\label{sec:results}

\begin{table}[t]
\centering
\caption{Exact-match accuracy in percent, mean $\pm$ standard deviation over $4$ training seeds, for the three setups at the four novelty levels. L1 is measured on the \splitmixed{} split and L0 on the held-out puzzles of the validation worlds. L2 is measured on draw $1$ of the no-mix split, on worlds containing cross-group pairs. Both L3 columns use the L3 dial split with the same held-out instances and test worlds: unsupported at level $0$ (no supporting instance in training), supported at the natural exposure level ($228$ supporting worlds). Best value per column in bold.}
\label{tab:main-results}
\footnotesize
\setlength{\tabcolsep}{5pt}
\begin{tabular}{@{}l r r r r r@{}}
\toprule
 & \multicolumn{2}{c}{\splitmixed{} split} & no-mix, draw 1 & \multicolumn{2}{c}{L3 dial split} \\
\cmidrule(lr){2-3}\cmidrule(lr){4-4}\cmidrule(lr){5-6}
Setup & L0 & L1 & L2 & L3 unsupported & L3 supported \\
\midrule
\noinfo{}    & $60.5 \pm 0.8$ & $65.4 \pm 1.6$ & $63.1 \pm 1.8$ & $52.1 \pm 2.3$ & $56.7$ \\
\descsetup{} & $\mathbf{99.2 \pm 0.1}$ & $\mathbf{98.5 \pm 0.2}$ & $67.6 \pm 5.5$ & $34.6 \pm 6.4$ & $79.0 \pm 5.4$ \\
\exsetup{}   & $95.2 \pm 1.2$ & $95.4 \pm 1.1$ & $\mathbf{82.3 \pm 1.6}$ & $\mathbf{65.0 \pm 0.8}$ & $\mathbf{90.8 \pm 2.3}$ \\
\bottomrule
\end{tabular}
\end{table}

We compare the three setups of Table~\ref{tab:setups} at the four novelty levels of Section~\ref{sec:testing-different-modes-of-novelty}, with one model and one training recipe throughout, and place a \noinfo{} floor beside every number (Table~\ref{tab:main-results}). In one sentence: the \descsetup{} setup matches or exceeds the \exsetup{} setup at recombining trained rules, and falls below the \noinfo{} floor on rule instances whose components were never seen together in training.

\paragraph{Recombining trained rules (L0 to L2).}
On the \splitmixed{} split the \descsetup{} setup reaches $99.2\%$ (L0) and $98.5\%$ (L1), the \exsetup{} setup $95.2\%$ and $95.4\%$, and the \noinfo{} floor $60.5\%$ and $65.4\%$. Under the training budget used, the description is thus not worse than examples at recombining trained rules (here it is even $3$ to $4$ percentage points better). This ranking depends on the training budget: a shorter schedule reverses it (Appendix~\ref{app:provenance}). Both setups fit their training puzzles, so neither is capacity-limited. When the training set never crosses the boundary between the two rule groups (L2), both setups fall far below their L1 accuracy: the description to $67.6 \pm 5.5\%$ and examples to $82.3 \pm 1.6\%$. Therefore, rule co-occurrence matters to both setups and the \descsetup{} setup depends on it more.

\paragraph{New rule instances (L3).}
When the held-out instances are supported (every pair of their components occurs in some trained rule), both setups are far above the \noinfo{} floor. When they are unsupported, the two setups separate on opposite sides of it (Table~\ref{tab:main-results}): the \descsetup{} setup drops to $34.6\%$, $17.5$ points \emph{below} the \noinfo{} model, while the \exsetup{} setup stays $13$ points above it. This pattern replicates on six further draws of two other held-out sets: on every draw the description lands $9$ to $17$ points below the floor and examples $8$ to $15$ points above it (Appendix~\ref{app:results-splits}). The deficit is not explained by training length, by the amount of training data, or by how the description is injected into the model (Appendix~\ref{app:controls}).

\paragraph{Exposure to compositional structure.}
Both dials add co-occurring structure to the training set while the test worlds stay fixed (Figure~\ref{fig:exposure}, Appendix~\ref{app:exposure}). On the L2 dial, as training worlds that mix the two rule groups are added, both setups climb toward their L1 accuracy while the \noinfo{} floor rises only from $62$ to $70$. The description starts $15$ points behind examples and overtakes them once about a fifth of the training worlds mix the groups. On the L3 dial, with identical held-out instances at every level, the description starts $17.5$ points below the \noinfo{} floor and ends $22$ points above it; examples rise too, but by $26$ points against $44$ for the description. On the splits tested, what limits symbolic composition is therefore whether the components of a new rule were seen together in training, not whether each component was seen by itself. By rule family, the description trails examples at the natural exposure level only on parity rules (Appendix~\ref{app:families}).

\begin{table}[t]
\centering
\caption{Description dropout. Exact match in percent, mean $\pm$ standard deviation over $4$ seeds, evaluated with the description present; floor is the \noinfo{} model on the same split. The last row shows the cost on in-support rules. Full results, including further splits and evaluation with the description masked, in Appendix~\ref{app:dropout}.}
\label{tab:dropout-main}
\footnotesize
\setlength{\tabcolsep}{5pt}
\begin{tabular}{@{}l r r r r@{}}
\toprule
Setting & \descsetup{} & + dropout & floor & \exsetup{} \\
\midrule
L3 dial, level 0 (unsupported) & $34.6 \pm 6.4$ & $58.8 \pm 1.6$ & $52.1 \pm 2.3$ & $65.0 \pm 0.8$ \\
no-mix split, L2               & $67.6 \pm 5.5$ & $71.8 \pm 2.2$ & $63.1 \pm 1.8$ & $82.3 \pm 1.6$ \\
no-mix split, L3               & $58.7 \pm 3.2$ & $66.6 \pm 2.1$ & $61.7 \pm 1.6$ & $77.0 \pm 2.2$ \\
L3 dial, level 0, L1 (in support) & $98.3 \pm 0.9$ & $96.5 \pm 0.2$ & $75.4 \pm 0.9$ & $98.0 \pm 0.1$ \\
\bottomrule
\end{tabular}
\end{table}
 
\paragraph{Why the description hurts.}
The pattern is consistent with the description acting as a crutch: a model that receives the correct description on every training puzzle never needs to infer the rules from the grid, so when the description names a pair of components it has never composed, it has nothing to fall back on, whereas the \noinfo{} model, forced to infer, does better. Removing the crutch supports this: replacing the description by null tokens on half of the training samples (description dropout) lifts the \descsetup{} model above the floor on every setting tried, at a cost of $1$ to $2$ points on L1 (Table~\ref{tab:dropout-main}; Appendix~\ref{app:dropout}).
 
\section{Related work}
\label{sec:related-work}

\paragraph{Formal definitions of intelligence.}
Legg and Hutter~\citep{legg2007universal} and Hern\'{a}ndez-Orallo~\citep{hernandez2017measure} formalize intelligence as aggregate task performance but specify no canonical, computable task family with explicit difficulty; our parameterized family is a concrete instance, also lining up with Chollet's local/broad spectrum~\citep{chollet2019arc} (extended discussion in Appendix~\ref{app:discussion}).

\paragraph{Rule inference.}
ARC~\citep{chollet2019arc}, MiniGrid/BabyAI~\citep{chevalier2018babyai}, CLEVR~\citep{johnson2017clevr}, and Transformer CSP-solvers~\citep{shah2024causal,giannoulis2025trial} evaluate rule inference but remain shaped by human-designed vocabulary or a fixed problem family, unlike our combinatorial rule vocabulary. DiG-bench~\citep{battleday2026digbench} also benchmarks rule discovery, but through long, text-based game transcripts that must be read as large context, a setting not adapted to the implicit, small-recursive-model style of reasoning our setup targets. Every world here is also a structured CSP (Section~\ref{sec:puzzles-def}) whose regular geometry (Appendix~\ref{app:pool6}) is what makes pool rules decompose into the type/axis/position/parameter factors our \descsetup{} setup exploits.

\paragraph{Compositional generalization.}
SCAN~\citep{lake2018generalization} and COGS~\citep{kim2020cogs} test a closely related property - generalizing to unseen combinations of familiar components - and find large in-distribution/compositional gaps. Our composition is instead over several constraints that must hold \emph{simultaneously} on a shared grid rather than sequential linguistic structure, and our \texttt{L3} tier withholds a rule instance from training \emph{entirely}, stricter than SCAN's primitive split, where the held-out primitive is still demonstrated alone during training. Most closely related, \citet{dziri2023faithandfate} find compositional generalization on multiplication/logic-grid puzzles collapses with \emph{computation-graph} depth (\emph{linearized subgraph matching} rather than a general algorithm); their complexity axis is instance \emph{size}, ours \emph{which rules govern the instance}. \citet{schug2024modular} show that modular solutions that generalize compositionally require a training distribution with connected support over modules; our no-mix split violates that condition by construction, and our exposure dials measure how much co-occurrence beyond it is needed, separately for symbolic descriptions and for demonstrations.

\section{Limitations and further research}
\label{sec:limitations}

\begin{itemize}[itemsep=1pt,topsep=2pt,parsep=1pt,partopsep=0pt,leftmargin=*]
  \item \textbf{Exploring other applications of the framework.} We instantiate the abstract framework of Section~\ref{sec:worlds-framework} with one concrete family of Sudoku-style grid worlds, but nothing in the formalism is specific to this choice - other parameterized families, over other rule vocabularies and grid structures, could be built the same way. The construction also extends naturally to space-time trajectories of a discrete dynamical system by adding a time axis, so that forbidden patterns span both space and time; exploring such dynamical instances is left to future work.
  \item \textbf{Only one family of implicit reasoning models tested.} All experiments use one HRM-style architecture (Section~\ref{sec:model}); we have not tested other variants of implicit, recurrent reasoning models along the same rule-adaptation dimension studied here, nor compared against large language models, models using explicit reasoning, or agentic frameworks, which could in principle approach these tasks very differently. Whether the conclusions of Section~\ref{sec:results} transfer to these other classes of models is untested.
  \item \textbf{Variations of the design choices.} Several of our design choices are held fixed across every world and setup even though the right value could plausibly vary with the specific rule combination; we did not systematically explore this. For example, the \exsetup{} setup always shows $M=2$ worked examples, a default carried over from an earlier version of the design rather than derived from a per-world or per-rule-type identifiability argument. Earlier experimentation varying $M$ suggested that how many examples are actually needed to disambiguate a world's rule combination depends on the world itself, so a single fixed $M$ may be more than necessary for some worlds and too few for others; characterizing this per-world requirement is left to future work.
\end{itemize}
Appendix~\ref{app:limitations} lists further limitations of the evidence.
 
\section{Conclusion}

We proposed a framework for studying adaptation to new worlds in a controlled way, where a world is defined as a finite grid with a set of forbidden patterns. We create a family of worlds by choosing which rules from a fixed pool are active. At test time, we can control the type of novelty: giving the model a new puzzle in a familiar world, a new combination of trained rules, a pairing of rules that were never seen together, or a rule that was never seen in training.
Because the rules decompose into reusable components, the framework also lets one control how much co-occurring structure the training set exposes while holding the test worlds fixed, and lets one compare ways of communicating the active rules (a symbolic description or solved examples) under identical conditions. The formalism is not tied to the Sudoku-like family instantiated here. Other rule vocabularies, grid structures, and, with a time axis, discrete dynamical systems fit the same definitions.

As a first use of the framework, we trained one depth-recurrent model across many combinations of rules and measured its accuracy at the four novelty levels. The experiments showed that recombining trained rules is easy whichever way the rules are communicated. Applying a rule whose components were never seen together is hard for both, and here the symbolic description does worse than no description at all. Adding co-occurring structure to the training set, or replacing the description by null tokens on half of the training samples (description dropout), repairs this. 
To adapt to a new world, a model needs to see not just each component of a new rule on its own, but also see those components together. Our framework makes it possible to measure and control how much of this combined information the model sees in training.
 
\ifdefined\arxivversion\section*{Acknowledgements}
This work was supported by the project ``Biography of Fake News with a Touch of AI: Dangerous Phenomenon through the Prism of Modern Human Sciences'' (Project No. CZ.02.01.01/00/23\_025/0008724) via the Operational Programme Jan \'{A}mos Komensk\'{y}. We acknowledge the EuroHPC Joint Undertaking for awarding the project ID EHPC-BEN-2026B07-121, ``Learning the causal structure of two-dimensional subshifts of finite type at scale'' (PI Jan H\r{u}la, Czech Technical University), access to the LEONARDO supercomputer, hosted by CINECA (Italy), through the EuroHPC JU Benchmark Access call.
 \fi

\section*{AI Usage Statement}
AI tools were used throughout the research and writing process as collaborative assistants. The main research ideas and conceptual motivation originated from the authors. AI was subsequently used to help operationalize these ideas, develop and execute parts of the experimental workflow, and draft portions of the manuscript.

In particular, Claude Code was used iteratively to assist with implementation and experimentation. The authors continuously reviewed, discussed, and evaluated the code and proposed approaches, providing feedback and refining the direction of the work throughout this process. Thus, the use of AI involved an iterative feedback loop between the authors and the AI system instead of autonomous generation of the research.

AI tools were also used to assist in drafting sections of the paper. Drafts were repeatedly reviewed and revised by the authors, who corrected inaccurate or misleading statements, requested modifications, incorporated additional ideas, and determined the final content and interpretation. Code, experimental procedures, results, mathematical statements, and citations suggested or generated with the assistance of AI were checked by the authors before being incorporated into the work.

The authors are responsible for the final methodology, claims, analysis, interpretation, and presentation of the work.
 
\bibliographystyle{iclr2027_conference}
\bibliography{biblio}

\appendix

\section*{Appendix}

This appendix is organized into self-contained parts, each expanding on one aspect of the main text: Appendix~\ref{app:pool6} lists every rule in the pool; Appendix~\ref{app:generation} details how puzzles are generated and solved; Appendix~\ref{app:results} gives the exposure dials, the further held-out sets, the description-dropout results, all controls, the provenance of an earlier result, the breakdown by rule family, and further limitations of the evidence; Appendix~\ref{app:discussion} expands on two discussions of the main text; Appendix~\ref{app:model} gives the full architecture and training specification; and Appendix~\ref{app:design-choices} gives the rationale behind task and data design choices. Each part can be read on its own.
 
\section{Full rule-pool definitions}
\label{app:pool6}

This appendix lists every rule in $\mathcal{V}$, the fixed $49$-instance pool from which each world's $2$--$4$ variable rules are drawn (Section~\ref{sec:family}), grouped by the three families already introduced in Table~\ref{tab:pool6-families}.

Recall that the grid is $6\times6$, with symbols $\{1,\ldots,6\}$. Every puzzle is always subject to the base Latin-square constraint (each row and each column contains all six symbols exactly once); on top of this base, each specific world adds $2$--$4$ extra rules drawn from $\mathcal{V}$.

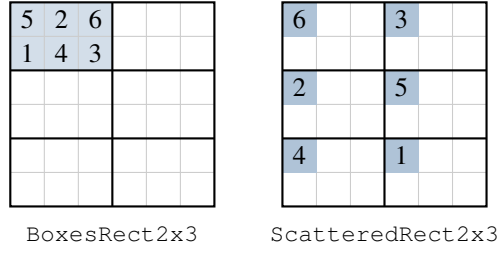
\begin{figure}[h]
\centering
\begin{tikzpicture}[scale=0.45]
\begin{scope}
    \foreach \x in {0,...,6} \draw[gray!40] (\x,0) -- (\x,6);
    \foreach \y in {0,...,6} \draw[gray!40] (0,\y) -- (6,\y);
    \fill[cRegion!25] (0,4) rectangle (3,6);
    \draw[gray!40] (1,4) -- (1,6);
    \draw[gray!40] (2,4) -- (2,6);
    \draw[gray!40] (0,5) -- (3,5);
    \draw[thick] (0,0) rectangle (6,6);
    \draw[thick] (3,0) -- (3,6);
    \draw[thick] (0,2) -- (6,2);
    \draw[thick] (0,4) -- (6,4);
    \node at (0.5,5.5) {5};
    \node at (1.5,5.5) {2};
    \node at (2.5,5.5) {6};
    \node at (0.5,4.5) {1};
    \node at (1.5,4.5) {4};
    \node at (2.5,4.5) {3};
    \node at (3,-0.8) {\footnotesize \texttt{BoxesRect2x3}};
  \end{scope}
\begin{scope}[shift={(8,0)}]
    \foreach \x in {0,...,6} \draw[gray!40] (\x,0) -- (\x,6);
    \foreach \y in {0,...,6} \draw[gray!40] (0,\y) -- (6,\y);
    \fill[cRegion!45] (0,5) rectangle (1,6);
    \fill[cRegion!45] (3,5) rectangle (4,6);
    \fill[cRegion!45] (0,3) rectangle (1,4);
    \fill[cRegion!45] (3,3) rectangle (4,4);
    \fill[cRegion!45] (0,1) rectangle (1,2);
    \fill[cRegion!45] (3,1) rectangle (4,2);
    \draw[thick] (0,0) rectangle (6,6);
    \draw[thick] (3,0) -- (3,6);
    \draw[thick] (0,2) -- (6,2);
    \draw[thick] (0,4) -- (6,4);
    \node at (0.5,5.5) {6};
    \node at (3.5,5.5) {3};
    \node at (0.5,3.5) {2};
    \node at (3.5,3.5) {5};
    \node at (0.5,1.5) {4};
    \node at (3.5,1.5) {1};
    \node at (3,-0.8) {\footnotesize \texttt{ScatteredRect2x3}};
  \end{scope}
\end{tikzpicture}
\caption{Left: the \texttt{BoxesRect2x3} rule requires the six cells physically inside one $2\times3$ box (shaded, one possible filling shown) to together contain every symbol exactly once. Right: its scattered counterpart \texttt{ScatteredRect2x3} instead groups the six cells that share the same position \emph{across} the grid's six boxes (here, each box's top-left corner, shaded) into one such group.}
\label{fig:box-scattered}
\end{figure}

\begin{itemize}
  \item \textbf{Box variants:} \texttt{BoxesRect2x3}, \texttt{BoxesRect3x2}: the grid is cut into $2\times3$ or $3\times2$ boxes, and each box must contain every symbol exactly once (a rectangular generalization of the classic $3\times3$ Sudoku box rule). Their scattered counterparts, \texttt{ScatteredRect2x3}/\texttt{ScatteredRect3x2}, group cells across boxes instead of within them: for each of the $6$ positions a cell can occupy inside a box (e.g.\ its top-left corner), the $6$ cells at that position - one from each of the grid's $6$ boxes - must together contain every symbol exactly once. Figure~\ref{fig:box-scattered} illustrates both.
\end{itemize}

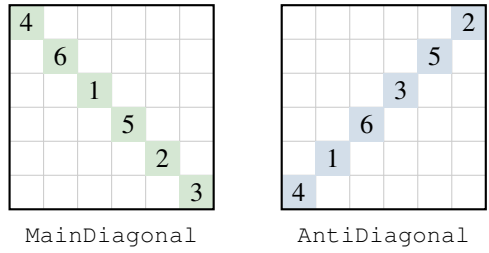
\begin{figure}[h]
\centering
\begin{tikzpicture}[scale=0.45]
\begin{scope}
    \foreach \x in {0,...,6} \draw[gray!40] (\x,0) -- (\x,6);
    \foreach \y in {0,...,6} \draw[gray!40] (0,\y) -- (6,\y);
    \fill[cParity!25] (0,5) rectangle (1,6);
    \fill[cParity!25] (1,4) rectangle (2,5);
    \fill[cParity!25] (2,3) rectangle (3,4);
    \fill[cParity!25] (3,2) rectangle (4,3);
    \fill[cParity!25] (4,1) rectangle (5,2);
    \fill[cParity!25] (5,0) rectangle (6,1);
    \draw[thick] (0,0) rectangle (6,6);
    \node at (0.5,5.5) {4};
    \node at (1.5,4.5) {6};
    \node at (2.5,3.5) {1};
    \node at (3.5,2.5) {5};
    \node at (4.5,1.5) {2};
    \node at (5.5,0.5) {3};
    \node at (3,-0.8) {\footnotesize \texttt{MainDiagonal}};
  \end{scope}
\begin{scope}[shift={(8,0)}]
    \foreach \x in {0,...,6} \draw[gray!40] (\x,0) -- (\x,6);
    \foreach \y in {0,...,6} \draw[gray!40] (0,\y) -- (6,\y);
    \fill[cRegion!25] (5,5) rectangle (6,6);
    \fill[cRegion!25] (4,4) rectangle (5,5);
    \fill[cRegion!25] (3,3) rectangle (4,4);
    \fill[cRegion!25] (2,2) rectangle (3,3);
    \fill[cRegion!25] (1,1) rectangle (2,2);
    \fill[cRegion!25] (0,0) rectangle (1,1);
    \draw[thick] (0,0) rectangle (6,6);
    \node at (5.5,5.5) {2};
    \node at (4.5,4.5) {5};
    \node at (3.5,3.5) {3};
    \node at (2.5,2.5) {6};
    \node at (1.5,1.5) {1};
    \node at (0.5,0.5) {4};
    \node at (3,-0.8) {\footnotesize \texttt{AntiDiagonal}};
  \end{scope}
\end{tikzpicture}
\caption{Left: \texttt{MainDiagonal} requires the six cells on the $\searrow$ diagonal (shaded, one possible filling shown) to together contain every symbol exactly once. Right: \texttt{AntiDiagonal} imposes the same requirement on the six cells of the $\swarrow$ diagonal instead.}
\label{fig:diagonals}
\end{figure}

\begin{itemize}
  \item \textbf{Diagonal variants:} \texttt{MainDiagonal} (the $\searrow$ diagonal) and \texttt{AntiDiagonal} (the $\swarrow$ diagonal) each independently require a unique occurrence of each symbol across their six cells, kept as separate rule instances rather than bundled into one (one of several design choices that makes this pool more \emph{factorizable} into reusable, independently-composable pieces). Figure~\ref{fig:diagonals} illustrates both.
\end{itemize}

\begin{figure}[h]
\centering
\begin{tikzpicture}[scale=0.45]
\begin{scope}
    \foreach \x in {0,...,6} \draw[gray!40] (\x,0) -- (\x,6);
    \foreach \y in {0,...,6} \draw[gray!40] (0,\y) -- (6,\y);
    \draw[thick] (0,0) rectangle (6,6);
    \fill[cAdjCell!45] (2,3) rectangle (3,4);
    \fill[cAdj!25] (1,5) rectangle (2,6);
    \fill[cAdj!25] (3,5) rectangle (4,6);
    \fill[cAdj!25] (0,4) rectangle (1,5);
    \fill[cAdj!25] (4,4) rectangle (5,5);
    \fill[cAdj!25] (0,2) rectangle (1,3);
    \fill[cAdj!25] (4,2) rectangle (5,3);
    \fill[cAdj!25] (1,1) rectangle (2,2);
    \fill[cAdj!25] (3,1) rectangle (4,2);
    \draw[thick] (0,0) rectangle (6,6);
    \node at (3,-0.8) {\footnotesize \texttt{AntiKnight}};
  \end{scope}
\begin{scope}[shift={(8,0)}]
    \foreach \x in {0,...,6} \draw[gray!40] (\x,0) -- (\x,6);
    \foreach \y in {0,...,6} \draw[gray!40] (0,\y) -- (6,\y);
    \draw[thick] (0,0) rectangle (6,6);
    \fill[cAdjCell!45] (2,3) rectangle (3,4);
    \fill[cAdj!25] (1,4) rectangle (2,5);
    \fill[cAdj!25] (3,4) rectangle (4,5);
    \fill[cAdj!25] (1,2) rectangle (2,3);
    \fill[cAdj!25] (3,2) rectangle (4,3);
    \draw[thick] (0,0) rectangle (6,6);
    \node at (3,-0.8) {\footnotesize \texttt{AntiKing}};
  \end{scope}
\end{tikzpicture}
\caption{Left: \texttt{AntiKnight} forbids the highlighted cell (dark grey) from sharing a value with any cell a chess knight's-move away (red). Right: \texttt{AntiKing} forbids the same cell from sharing a value with any diagonally-adjacent cell (red).}
\label{fig:knight-king}
\end{figure}
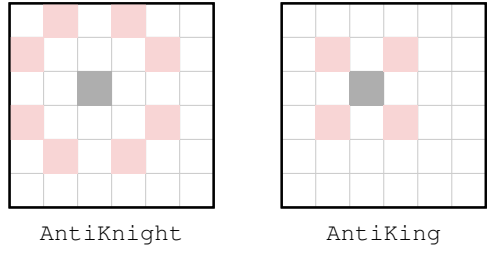

\begin{itemize}
  \item \textbf{Adjacency constraints:} \texttt{AntiKnight} forbids any two cells a chess knight's-move apart from sharing a value; \texttt{AntiKing} forbids any two diagonally-adjacent cells from sharing a value. Figure~\ref{fig:knight-king} illustrates both.
\end{itemize}

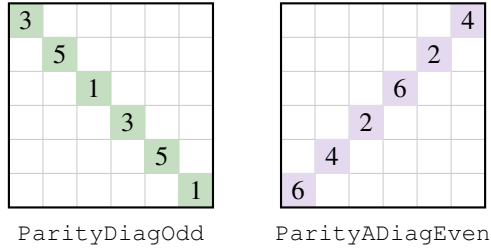
\begin{figure}[h]
\centering
\begin{tikzpicture}[scale=0.45]
\begin{scope}
    \foreach \x in {0,...,6} \draw[gray!40] (\x,0) -- (\x,6);
    \foreach \y in {0,...,6} \draw[gray!40] (0,\y) -- (6,\y);
    \fill[cParity!35] (0,5) rectangle (1,6);
    \fill[cParity!35] (1,4) rectangle (2,5);
    \fill[cParity!35] (2,3) rectangle (3,4);
    \fill[cParity!35] (3,2) rectangle (4,3);
    \fill[cParity!35] (4,1) rectangle (5,2);
    \fill[cParity!35] (5,0) rectangle (6,1);
    \draw[thick] (0,0) rectangle (6,6);
    \node at (0.5,5.5) {3};
    \node at (1.5,4.5) {5};
    \node at (2.5,3.5) {1};
    \node at (3.5,2.5) {3};
    \node at (4.5,1.5) {5};
    \node at (5.5,0.5) {1};
    \node at (3,-0.8) {\footnotesize \texttt{ParityDiagOdd}};
  \end{scope}
\begin{scope}[shift={(8,0)}]
    \foreach \x in {0,...,6} \draw[gray!40] (\x,0) -- (\x,6);
    \foreach \y in {0,...,6} \draw[gray!40] (0,\y) -- (6,\y);
    \fill[cAlt1!25] (5,5) rectangle (6,6);
    \fill[cAlt1!25] (4,4) rectangle (5,5);
    \fill[cAlt1!25] (3,3) rectangle (4,4);
    \fill[cAlt1!25] (2,2) rectangle (3,3);
    \fill[cAlt1!25] (1,1) rectangle (2,2);
    \fill[cAlt1!25] (0,0) rectangle (1,1);
    \draw[thick] (0,0) rectangle (6,6);
    \node at (5.5,5.5) {4};
    \node at (4.5,4.5) {2};
    \node at (3.5,3.5) {6};
    \node at (2.5,2.5) {2};
    \node at (1.5,1.5) {4};
    \node at (0.5,0.5) {6};
    \node at (3,-0.8) {\footnotesize \texttt{ParityADiagEven}};
  \end{scope}
\end{tikzpicture}
\caption{Left: \texttt{ParityDiagOdd} restricts every cell on the main diagonal (shaded) to the odd values $\{1,3,5\}$; unlike the earlier diagonal rule, repeats are allowed here, since membership in this fixed set is all that is required. Right: \texttt{ParityADiagEven} applies the analogous restriction to the anti-diagonal, with the even set $\{2,4,6\}$. The remaining two combinations, \texttt{ParityDiagEven} and \texttt{ParityADiagOdd}, swap which parity set applies to which diagonal.}
\label{fig:parity-diag}
\end{figure}

\begin{itemize}
  \item \textbf{Parity-on-diagonal rules:} \texttt{ParityDiagOdd}/\texttt{ParityDiagEven} restrict every cell on the main diagonal to only odd values $\{1,3,5\}$ or only even values $\{2,4,6\}$, respectively; \texttt{ParityADiagOdd}/\texttt{ParityADiagEven} apply the same restriction to the anti-diagonal instead. Figure~\ref{fig:parity-diag} illustrates two of the four combinations.
\end{itemize}

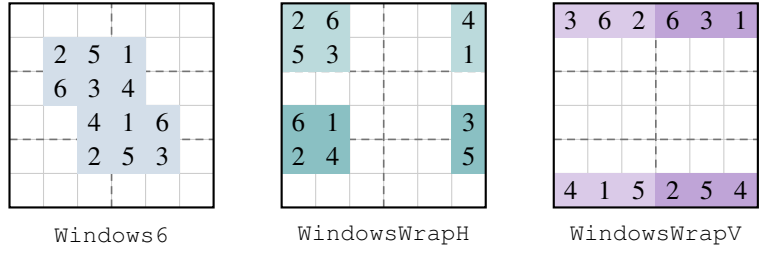
\begin{figure}[h]
\centering
\begin{tikzpicture}[scale=0.45]
\begin{scope}
    \foreach \x in {0,...,6} \draw[gray!40] (\x,0) -- (\x,6);
    \foreach \y in {0,...,6} \draw[gray!40] (0,\y) -- (6,\y);
    \draw[black!55,densely dashed,line width=0.6pt] (3,0) -- (3,6);
    \draw[black!55,densely dashed,line width=0.6pt] (0,2) -- (6,2);
    \draw[black!55,densely dashed,line width=0.6pt] (0,4) -- (6,4);
    \fill[cRegion!25] (1,3) rectangle (4,5);
    \fill[cRegion!25] (2,1) rectangle (5,3);
    \draw[thick] (0,0) rectangle (6,6);
    \node at (1.5,4.5) {2}; \node at (2.5,4.5) {5}; \node at (3.5,4.5) {1};
    \node at (1.5,3.5) {6}; \node at (2.5,3.5) {3}; \node at (3.5,3.5) {4};
    \node at (2.5,2.5) {4}; \node at (3.5,2.5) {1}; \node at (4.5,2.5) {6};
    \node at (2.5,1.5) {2}; \node at (3.5,1.5) {5}; \node at (4.5,1.5) {3};
    \node at (3,-0.8) {\footnotesize \texttt{Windows6}};
  \end{scope}
\begin{scope}[shift={(8,0)}]
    \foreach \x in {0,...,6} \draw[gray!40] (\x,0) -- (\x,6);
    \foreach \y in {0,...,6} \draw[gray!40] (0,\y) -- (6,\y);
    \draw[black!55,densely dashed,line width=0.6pt] (3,0) -- (3,6);
    \draw[black!55,densely dashed,line width=0.6pt] (0,2) -- (6,2);
    \draw[black!55,densely dashed,line width=0.6pt] (0,4) -- (6,4);
    \fill[cAlt2!35] (0,4) rectangle (2,6);
    \fill[cAlt2!35] (5,4) rectangle (6,6);
    \fill[cAlt2!60] (0,1) rectangle (2,3);
    \fill[cAlt2!60] (5,1) rectangle (6,3);
    \draw[thick] (0,0) rectangle (6,6);
    \node at (0.5,5.5) {2}; \node at (1.5,5.5) {6}; \node at (5.5,5.5) {4};
    \node at (0.5,4.5) {5}; \node at (1.5,4.5) {3}; \node at (5.5,4.5) {1};
    \node at (0.5,2.5) {6}; \node at (1.5,2.5) {1}; \node at (5.5,2.5) {3};
    \node at (0.5,1.5) {2}; \node at (1.5,1.5) {4}; \node at (5.5,1.5) {5};
    \node at (3,-0.8) {\footnotesize \texttt{WindowsWrapH}};
  \end{scope}
\begin{scope}[shift={(16,0)}]
    \foreach \x in {0,...,6} \draw[gray!40] (\x,0) -- (\x,6);
    \foreach \y in {0,...,6} \draw[gray!40] (0,\y) -- (6,\y);
    \draw[black!55,densely dashed,line width=0.6pt] (3,0) -- (3,6);
    \draw[black!55,densely dashed,line width=0.6pt] (0,2) -- (6,2);
    \draw[black!55,densely dashed,line width=0.6pt] (0,4) -- (6,4);
    \fill[cAlt1!35] (0,5) rectangle (3,6);
    \fill[cAlt1!35] (0,0) rectangle (3,1);
    \fill[cAlt1!60] (3,5) rectangle (6,6);
    \fill[cAlt1!60] (3,0) rectangle (6,1);
    \draw[thick] (0,0) rectangle (6,6);
    \node at (0.5,5.5) {3}; \node at (1.5,5.5) {6}; \node at (2.5,5.5) {2};
    \node at (3.5,5.5) {6}; \node at (4.5,5.5) {3}; \node at (5.5,5.5) {1};
    \node at (0.5,0.5) {4}; \node at (1.5,0.5) {1}; \node at (2.5,0.5) {5};
    \node at (3.5,0.5) {2}; \node at (4.5,0.5) {5}; \node at (5.5,0.5) {4};
    \node at (3,-0.8) {\footnotesize \texttt{WindowsWrapV}};
  \end{scope}
\end{tikzpicture}
\caption{The three window placements (shaded, one possible filling shown), with the $2\times3$ box tiling from Figure~\ref{fig:box-scattered} overlaid as a dashed reference grid. \texttt{Windows6}'s pair straddles the box boundaries, in the spirit of a real Windoku puzzle. \texttt{WindowsWrapH} and \texttt{WindowsWrapV} instead each wrap a window pair around the grid's border (matching shade intensity marks the two pieces of the same window, split across the wrap); neither ever coincides with a \texttt{BoxesRect2x3} box.}
\label{fig:windows}
\end{figure}

\begin{itemize}
  \item \textbf{Windows:} \texttt{Windows6} places a pair of extra $2\times3$ ``window'' regions at a fixed pair of corner offsets in the grid, each independently a unique-occurrence group, offset from the \texttt{BoxesRect2x3} box tiling so that it straddles box boundaries, in the spirit of a real Windoku puzzle. \texttt{WindowsWrapH} and \texttt{WindowsWrapV} extend this idea by wrapping a window pair around the grid's border: \texttt{WindowsWrapH} takes a $3$-column band that wraps from the last column back to the first (columns $6,1,2$), instantiated at two different row positions; \texttt{WindowsWrapV} takes a $2$-row band that wraps from the last row back to the first (rows $6,1$), instantiated at two different column positions. Figure~\ref{fig:windows} shows all three, with the box tiling overlaid for reference.
\end{itemize}

\begin{figure}[h]
\centering
\begin{tikzpicture}[scale=0.45]
\begin{scope}
    \fill[cShift!15] (0,3) rectangle (6,6);
    \fill[cShift!35] (0,0) rectangle (6,3);
    \foreach \x in {0,...,6} \draw[gray!40] (\x,0) -- (\x,6);
    \foreach \y in {0,...,6} \draw[gray!40] (0,\y) -- (6,\y);
    \draw[thick] (0,0) rectangle (6,6);
    \draw[thick] (0,3) -- (6,3);
    \node at (0.5,5.5) {$a$}; \node at (0.5,4.5) {$b$}; \node at (0.5,3.5) {$c$};
    \node at (0.5,2.5) {$a'$}; \node at (0.5,1.5) {$b'$}; \node at (0.5,0.5) {$c'$};
    \draw[->,thick] (1,5.5) .. controls (2.6,5.5) and (2.6,2.5) .. (1,2.5);
    \node at (3.5,4) {\footnotesize $+k$ in the alphabet};
    \node at (3,-0.8) {\footnotesize \texttt{RowShiftK\{k\}}};
  \end{scope}
\begin{scope}[shift={(9,0)}]
    \fill[cShift!15] (0,0) rectangle (3,6);
    \fill[cShift!35] (3,0) rectangle (6,6);
    \foreach \x in {0,...,6} \draw[gray!40] (\x,0) -- (\x,6);
    \foreach \y in {0,...,6} \draw[gray!40] (0,\y) -- (6,\y);
    \draw[thick] (0,0) rectangle (6,6);
    \draw[thick] (3,0) -- (3,6);
    \node at (0.5,5.5) {$a$}; \node at (1.5,5.5) {$b$}; \node at (2.5,5.5) {$c$};
    \node at (3.5,5.5) {$a'$}; \node at (4.5,5.5) {$b'$}; \node at (5.5,5.5) {$c'$};
    \draw[->,thick] (0.5,6) .. controls (0.5,7.4) and (3.5,7.4) .. (3.5,6);
    \node at (2,7.7) {\footnotesize $+k$ in the alphabet};
    \node at (3,-0.8) {\footnotesize \texttt{ColShiftK\{k\}}};
  \end{scope}
\end{tikzpicture}
\caption{Both rules cut the grid into two fixed areas and then shift the \emph{alphabet value}, position by position, by $k$ between them - the areas themselves never move or depend on $k$ ($x'$ denotes $x$ shifted by $k$ in the alphabet, mod $6$). Left: \texttt{RowShiftK\{k\}} cuts the grid into a top and a bottom area (one column shown: top values $a,b,c$; the corresponding bottom values must be $a',b',c'$); every other column is cut and shifted the same way. Right: \texttt{ColShiftK\{k\}} is the column-wise analogue, cutting a left and a right area instead.}
\label{fig:shiftk}
\end{figure}
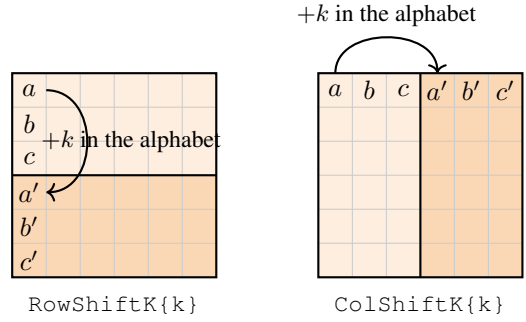

\begin{itemize}
  \item \textbf{Shift rules with explicit offset $k$:} \texttt{RowShiftK\{k\}} cuts the grid into two fixed areas, the top three rows and the bottom three rows, and requires that, position by position, the value in the bottom area equals the value in the top area shifted by $k$ in the alphabet (mod $6$) - the two areas themselves never change with $k$, only how far the alphabet value is shifted between them. \texttt{ColShiftK\{k\}} is the column-wise analogue, cutting the grid into a left and a right area instead. Figure~\ref{fig:shiftk} illustrates this for general $k$. In $\mathcal{V}$, only $k=3$ is used for these particular rules; $k=2$ and $k=4$ are provably unsatisfiable for this rule (informally: each base column's six linked cells must split evenly into a set $S$ and its shift $S+k$, which requires the shift map's orbits on $\{1,\ldots,6\}$ - of size $6/\gcd(k,6)$ - to be even; that size is the odd number $3$ exactly for $k\in\{2,4\}$, hence the conflict, while it is $6$ or $2$ for $k\in\{1,3,5\}$), and $k=5$, while satisfiable in isolation, admits no configuration in combination with any other pool rule, so neither was added.
\end{itemize}

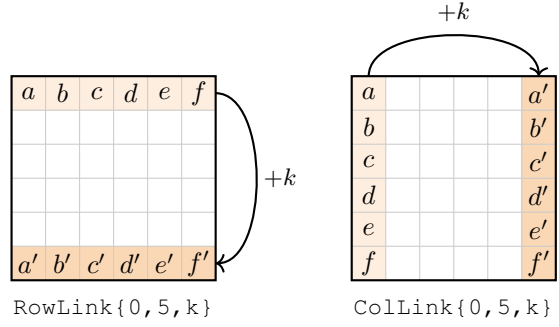
\begin{figure}[h]
\centering
\begin{tikzpicture}[scale=0.45]
\begin{scope}
    \fill[cShift!15] (0,5) rectangle (6,6);
    \fill[cShift!35] (0,0) rectangle (6,1);
    \foreach \x in {0,...,6} \draw[gray!40] (\x,0) -- (\x,6);
    \foreach \y in {0,...,6} \draw[gray!40] (0,\y) -- (6,\y);
    \draw[thick] (0,0) rectangle (6,6);
    \node at (0.5,5.5) {$a$}; \node at (1.5,5.5) {$b$}; \node at (2.5,5.5) {$c$};
    \node at (3.5,5.5) {$d$}; \node at (4.5,5.5) {$e$}; \node at (5.5,5.5) {$f$};
    \node at (0.5,0.5) {$a'$}; \node at (1.5,0.5) {$b'$}; \node at (2.5,0.5) {$c'$};
    \node at (3.5,0.5) {$d'$}; \node at (4.5,0.5) {$e'$}; \node at (5.5,0.5) {$f'$};
    \draw[->,thick] (6,5.5) .. controls (7.6,5.5) and (7.6,0.5) .. (6,0.5);
    \node at (7.9,3) {\footnotesize $+k$};
    \node at (3,-0.8) {\footnotesize \texttt{RowLink\{0,5,k\}}};
  \end{scope}
\begin{scope}[shift={(10,0)}]
    \fill[cShift!15] (0,0) rectangle (1,6);
    \fill[cShift!35] (5,0) rectangle (6,6);
    \foreach \x in {0,...,6} \draw[gray!40] (\x,0) -- (\x,6);
    \foreach \y in {0,...,6} \draw[gray!40] (0,\y) -- (6,\y);
    \draw[thick] (0,0) rectangle (6,6);
    \node at (0.5,5.5) {$a$}; \node at (0.5,4.5) {$b$}; \node at (0.5,3.5) {$c$};
    \node at (0.5,2.5) {$d$}; \node at (0.5,1.5) {$e$}; \node at (0.5,0.5) {$f$};
    \node at (5.5,5.5) {$a'$}; \node at (5.5,4.5) {$b'$}; \node at (5.5,3.5) {$c'$};
    \node at (5.5,2.5) {$d'$}; \node at (5.5,1.5) {$e'$}; \node at (5.5,0.5) {$f'$};
    \draw[->,thick] (0.5,6) .. controls (0.5,7.6) and (5.5,7.6) .. (5.5,6);
    \node at (3,7.9) {\footnotesize $+k$};
    \node at (3,-0.8) {\footnotesize \texttt{ColLink\{0,5,k\}}};
  \end{scope}
\end{tikzpicture}
\caption{Left: \texttt{RowLink\{0,5,k\}} links only row $1$ and row $6$ (shaded) - every column's value in row $6$ must equal its value in row $1$ shifted by $k$ in the alphabet ($x'$ denotes $x{+}k \bmod 6$); every other row is unconstrained by this rule. $\mathcal{V}$ includes two more such row pairs, $(2,5)$ and $(3,4)$ (1-indexed), and the column analogues, each at any $k\in\{1,\ldots,5\}$ as a separate instance. Right: \texttt{ColLink\{0,5,k\}} is the column-wise version, linking only column $1$ and column $6$.}
\label{fig:linkrules}
\end{figure}

\begin{itemize}
  \item \textbf{Link rules} ($30$ of the $49$ instances): \texttt{RowLink\{top,bottom,k\}} / \texttt{ColLink\{left,right,k\}} each link one specific pair of rows (or columns) with a value shift of $k$, cell by cell. $\mathcal{V}$ includes every such pair at $k\in\{1,2,3,4,5\}$; unlike the shift-rule family above, constraining only \emph{one} pair at a time avoids the orbit-parity conflict, so all four non-trivial $k$ values admit configurations here even though $k=2,4$ are unsatisfiable for \texttt{RowShiftK}/\texttt{ColShiftK}. This duplication is deliberate: each component (axis, pair of indices, offset $k$) appears in several contexts, so that a rule instance never seen as a whole can still be represented from its components. Figure~\ref{fig:linkrules} illustrates one row pair and one column pair.
\end{itemize}

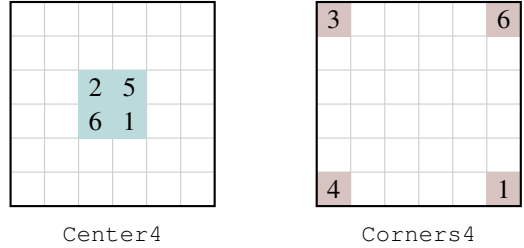
\begin{figure}[h]
\centering
\begin{tikzpicture}[scale=0.45]
\begin{scope}
    \foreach \x in {0,...,6} \draw[gray!40] (\x,0) -- (\x,6);
    \foreach \y in {0,...,6} \draw[gray!40] (0,\y) -- (6,\y);
    \fill[cAlt2!35] (2,2) rectangle (4,4);
    \draw[thick] (0,0) rectangle (6,6);
    \node at (2.5,3.5) {2}; \node at (3.5,3.5) {5};
    \node at (2.5,2.5) {6}; \node at (3.5,2.5) {1};
    \node at (3,-0.8) {\footnotesize \texttt{Center4}};
  \end{scope}
\begin{scope}[shift={(9,0)}]
    \foreach \x in {0,...,6} \draw[gray!40] (\x,0) -- (\x,6);
    \foreach \y in {0,...,6} \draw[gray!40] (0,\y) -- (6,\y);
    \fill[cAlt3!35] (0,5) rectangle (1,6);
    \fill[cAlt3!35] (5,5) rectangle (6,6);
    \fill[cAlt3!35] (0,0) rectangle (1,1);
    \fill[cAlt3!35] (5,0) rectangle (6,1);
    \draw[thick] (0,0) rectangle (6,6);
    \node at (0.5,5.5) {3}; \node at (5.5,5.5) {6};
    \node at (0.5,0.5) {4}; \node at (5.5,0.5) {1};
    \node at (3,-0.8) {\footnotesize \texttt{Corners4}};
  \end{scope}
\end{tikzpicture}
\caption{Left: \texttt{Center4} requires the four center cells (shaded) to all differ from each other - only $4$ of the $6$ symbols are involved, so unlike the region rules above, no symbol is required to appear. Right: \texttt{Corners4} imposes the same all-different requirement on the four corner cells instead.}
\label{fig:fixedregions}
\end{figure}

\begin{itemize}
  \item \textbf{Fixed small regions:} \texttt{Center4} requires the four center cells of the grid to all differ from each other; \texttt{Corners4} requires the same among the four corner cells. Neither requires every symbol to appear - with only $4$ cells and $6$ symbols, this is a plain all-different constraint, not a unique-occurrence region like the box/window/diagonal families above. Figure~\ref{fig:fixedregions} illustrates both.
\end{itemize}

The families above are not an exhaustive catalogue of \emph{natural} Sudoku-style constraints. For example, every rule here that relates two rows or columns (the shift and link families) relates their \emph{alphabet values} position by position, never their \emph{positions} themselves; a rule requiring row $6$ to equal row $1$ \emph{circularly shifted by some number of columns} - a positional, grid-wise rotation, rather than a value-wise one - would be an equally natural variant not present in $\mathcal{V}$.
 
\section{Puzzle generation and solver details}
\label{app:generation}

This appendix provides all details of puzzle-generation described succinctly in Section~\ref{sec:puzzles}. Figure~\ref{fig:example-configs} shows two admissible configurations of one concrete world, of the kind this pipeline samples from before digging a puzzle out of one of them.

\begin{figure}[t]
\centering
\begin{tikzpicture}[font=\footnotesize]
  \newcommand{\linkshade}{\fill[cShift!30] (0,5) rectangle (6,6); \fill[cShift!30] (0,0) rectangle (6,1);}
  \begin{scope}[scale=0.3]
    \linkshade\gridlines\boxlines
    \gridvalues{2,6,4,3,5,1, 3,1,5,4,6,2, 4,5,6,2,1,3, 1,2,3,5,4,6, 6,4,2,1,3,5, 5,3,1,6,2,4}
    \node[anchor=north] at (3,-0.3) {configuration 1};
  \end{scope}
  \begin{scope}[scale=0.3, shift={(8,0)}]
    \linkshade\gridlines\boxlines
    \gridvalues{3,2,5,1,4,6, 4,6,1,5,3,2, 2,1,6,3,5,4, 5,3,4,2,6,1, 1,4,3,6,2,5, 6,5,2,4,1,3}
    \node[anchor=north] at (3,-0.3) {configuration 2};
  \end{scope}
\end{tikzpicture}
\caption{Two admissible configurations of the world $\{\texttt{BoxesRect2x3},\ \texttt{RowLink(1,6,}k{=}3)\}$ on top of the Latin-square base (thick lines mark boxes; shaded rows are linked, row $6$ equals row $1$ shifted by $3$). These are the two solved grids the \exsetup{} setup (Section~\ref{sec:how-rules-communicated}) shows the model before a puzzle from this same world (Figure~\ref{fig:completion-problem} shows one such puzzle).}
\label{fig:example-configs}
\end{figure}
 
\paragraph{Non-emptiness check.} Not every combination of rules admits a configuration. For each candidate combination the generator samples full admissible configurations within a fixed time budget and digs a puzzle out of each one (see below). The combination becomes a world only if enough configurations were found and, in the median, at least $14$ of the $36$ cells could be blanked while keeping the solution unique, about the fraction blanked in a comparably hard $9\times9$ Sudoku. This is an empirical filter, not a proof of non-emptiness.

\paragraph{Generation pipeline.} Each puzzle is generated as follows:
\begin{enumerate}
  \item \textbf{Solving:} sample a valid configuration satisfying the world's constraints (base + extra rules) - a complete assignment with zero blanks (using the constraint-propagation and backtracking solver described below).
  \item \textbf{Digging:} starting from that full configuration, remove cells one at a time (setting them to blank), checking after each removal (using the solver again) that the remaining partial grid still has only one valid completion. If removing a cell would make the puzzle ambiguous (more than one grid fits the remaining clues), that removal is rejected and a different cell is tried.
  \item \textbf{Stopping at a target difficulty:} digging continues until the puzzle reaches a target number of blanks (a range calibrated per world by a separate search for a blank-count range that the world supports). More blanks generally means fewer given clues and a harder puzzle to solve, since more constraint-propagation is needed to pin down the unique answer. Reaching the target is not guaranteed: digging makes a single pass over the cells in one random order, and if that order is exhausted before the target is reached - because every remaining given cell is now needed for uniqueness - the entire attempt is discarded and generation restarts from a freshly sampled full configuration, rather than retrying the same one with a different cell order.
\end{enumerate}

\paragraph{Solver.} Solving and uniqueness checking use a standard constraint-propagation solver with backtracking: cell domains are tracked as bitmasks and shrunk by naked and hidden singles and by the pool's binary relations (not-equal, not-consecutive, equal-shifted-by-$k$); a minimum-remaining-values search resolves the rest, and uniqueness is established by counting solutions up to two. This is fast enough for $6\times6$ grids; larger grids would call for a SAT solver.

\paragraph{Dataset size per world.} We generate $2{,}000$ training and $200$ held-out puzzles per world (Section~\ref{sec:puzzles}).
 
\section{Additional results}
\label{app:results}

All numbers are exact-match accuracy in percent under the recipe of Section~\ref{sec:model} unless a table says ``screening recipe'' ($12$ epochs with cosine decay, used only for the early mechanism screen). ``Floor'' is the \noinfo{} model trained on the same split.

\subsection{Exposure dials}
\label{app:exposure}

Figure~\ref{fig:exposure} shows both dials of Section~\ref{sec:exposure}; Tables~\ref{tab:l2-dial} and~\ref{tab:l3-dial} give the numbers. In the L2 dial, $\rho$ is the fraction of training worlds containing at least one cross-group pair, and the pair coverage $c(\rho)$ is the fraction of cross-group pairs occurring in the test worlds that occur in at least one training world; the levels are $0$, $5$, $25$, $100$ and $373$ mixing worlds out of $500$, the last being the natural rate. In the L3 dial, five column links are held out at every level and the levels are $0$, $5$, $25$, $100$ and $228$ training worlds containing one of the ten remaining column links, the last again being the natural rate. In both dials the test worlds are identical at every level and the training sets are nested. The L1 columns of Table~\ref{tab:l3-dial} show that the runs whose L3 accuracy moves by $45$ points are tied at L1 at every level.

\begin{figure}[htb]
\centering
\pgfplotsset{
  exposure axis/.style={
    width=0.5\linewidth, height=5.2cm,
    ymajorgrids, grid style={black!10},
    tick label style={font=\scriptsize}, label style={font=\footnotesize},
    legend style={font=\scriptsize, draw=none, fill=none, legend columns=3, /tikz/every even column/.append style={column sep=6pt}},
    error bars/y dir=both, error bars/y explicit,
    error bars/error bar style={line width=0.4pt},
    every axis plot/.append style={line width=0.9pt, mark size=1.6pt},
    xtick=data, symbolic x coords={0,5,25,100,228,373},
  },
  desc/.style={color=cRegion, mark=*},
  ex/.style={color=cShift, mark=square*},
  floor/.style={color=black!55, mark=triangle*, dashed},
}
\begin{tikzpicture}
\begin{axis}[exposure axis, name=a,
  xlabel={mixing worlds (of 500); pair coverage $c(\rho)$ below},
  ylabel={L2 exact match (\%)}, ymin=55, ymax=100,
  xticklabels={0\\$c{=}0$,5\\$.05$,25\\$.34$,100\\$.64$,373\\$.86$},
  xticklabel style={align=center},
  title={(a) L2 dial}, title style={font=\footnotesize\bfseries}]
\addplot[desc]  coordinates {(0,67.6)+-(0,5.5) (5,70.0)+-(0,3.2) (25,89.2)+-(0,0.6) (100,96.6)+-(0,1.4) (373,98.5)+-(0,1.0)};
\addplot[ex]    coordinates {(0,82.3)+-(0,1.6) (5,88.6)+-(0,0.8) (25,92.6)+-(0,0.5) (100,95.8)+-(0,0.3) (373,97.0)+-(0,0.3)};
\addplot[floor] coordinates {(0,62.1) (5,63.8) (25,64.0) (100,67.7) (373,70.4)};
\end{axis}
\begin{axis}[exposure axis, at={(a.east)}, anchor=west, xshift=1.2cm,
  xlabel={supporting worlds (of 500)}, legend to name=exposurelegend,
  ylabel={L3 exact match (\%)}, ymin=25, ymax=95,
  xticklabels={0,5,25,100,228},
  title={(b) L3 dial}, title style={font=\footnotesize\bfseries}]
\addplot[desc]  coordinates {(0,34.6)+-(0,6.4) (5,48.5)+-(0,2.9) (25,48.0)+-(0,7.4) (100,79.0)+-(0,7.1) (228,79.0)+-(0,5.4)};
\addplot[ex]    coordinates {(0,65.0)+-(0,0.8) (5,70.0)+-(0,0.6) (25,77.0)+-(0,1.1) (100,86.6)+-(0,1.6) (228,90.8)+-(0,2.3)};
\addplot[floor] coordinates {(0,52.1) (5,52.5) (25,53.2) (100,57.7) (228,56.7)};
\addplot[color=cRegion, mark=diamond*, only marks, mark size=2.4pt] coordinates {(0,58.8)+-(0,1.6)};
\addplot[color=cShift, mark=square, only marks, mark size=1.8pt] coordinates {(0,68.2) (25,81.8) (228,91.9)};
\legend{{\descsetup{}}, {\exsetup{}}, {\noinfo{}}, {\descsetup{} + dropout}, {\exsetup{}, draw 2}}
\end{axis}
\end{tikzpicture}\\[2pt]
\pgfplotslegendfromname{exposurelegend}
\caption{Exposure dials (Section~\ref{sec:exposure}). (a) L2 accuracy on draw $1$ of the no-mix split as training worlds mixing the two rule groups are added; the test worlds are the same at every level and the training sets are nested. (b) L3 accuracy on the held-out column links as training worlds containing a supporting column link are added; held-out instances and test worlds are the same at every level. Points are means over $4$ seeds, bars are one standard deviation. The diamond is the \descsetup{} model trained with description dropout; hollow squares are the \exsetup{} setup on a second draw of the training worlds.}
\label{fig:exposure}
\end{figure}
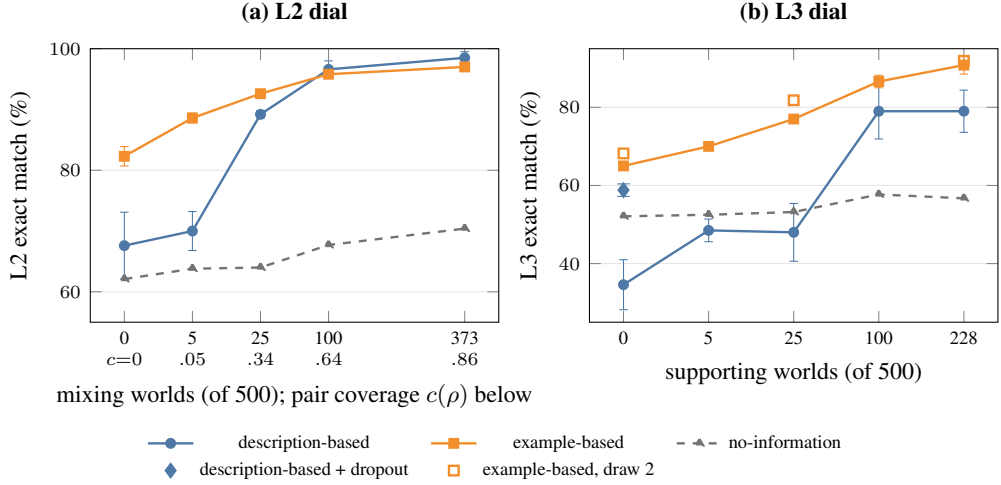
 \begin{table}[htb]
\centering
\caption{L2 exposure dial, draw $1$ of the no-mix split. Exact match in percent, mean $\pm$ standard deviation over $4$ seeds. Test sets are identical at every level; training sets are nested. L0 accuracy is $97.8$ to $99.5\%$ for the \descsetup{} and \exsetup{} setups at every level.}
\label{tab:l2-dial}
\footnotesize
\setlength{\tabcolsep}{4pt}
\begin{tabular}{@{}r r r r r r r r r@{}}
\toprule
mixing & $\rho$ & $c(\rho)$ & \multicolumn{3}{c}{L2} & \multicolumn{3}{c}{L3} \\
\cmidrule(lr){4-6}\cmidrule(lr){7-9}
worlds & & & \descsetup{} & \exsetup{} & floor & \descsetup{} & \exsetup{} & floor \\
\midrule
0   & 0     & 0.00 & $67.6 \pm 5.5$ & $82.3 \pm 1.6$ & 62.1 & $58.7 \pm 3.2$ & $77.0 \pm 2.2$ & 59.7 \\
5   & 0.01  & 0.05 & $70.0 \pm 3.2$ & $88.6 \pm 0.8$ & 63.8 & $63.6 \pm 4.8$ & $79.3 \pm 0.9$ & 61.9 \\
25  & 0.05  & 0.34 & $89.2 \pm 0.6$ & $92.6 \pm 0.5$ & 64.0 & $70.5 \pm 2.5$ & $78.8 \pm 1.4$ & 62.4 \\
100 & 0.20  & 0.64 & $96.6 \pm 1.4$ & $95.8 \pm 0.3$ & 67.7 & $73.4 \pm 0.8$ & $81.8 \pm 0.3$ & 61.2 \\
373 & 0.746 & 0.86 & $98.5 \pm 1.0$ & $97.0 \pm 0.3$ & 70.4 & $77.8 \pm 3.3$ & $82.0 \pm 1.4$ & 60.9 \\
\bottomrule
\end{tabular}
\end{table}
 \begin{table}[htb]
\centering
\caption{L3 exposure dial, draw $1$. Five column links are held out at every level; the level is the number of training worlds containing one of the ten remaining column links (supporting instances). Exact match in percent, mean $\pm$ standard deviation over $4$ seeds. Test sets are identical at every level. Second training draw at levels $0$, $25$ and $228$ ($2$ seeds): \exsetup{} $68.2 \pm 0.3$, $81.8 \pm 2.1$, $91.9 \pm 1.4$; \descsetup{} $39.8 \pm 0.2$, $69.2 \pm 1.8$, $84.4 \pm 3.3$.}
\label{tab:l3-dial}
\footnotesize
\setlength{\tabcolsep}{3pt}
\begin{tabular}{@{}r r r r r r r r@{}}
\toprule
support & supporting & \multicolumn{3}{c}{L3} & \multicolumn{3}{c}{L1} \\
\cmidrule(lr){3-5}\cmidrule(lr){6-8}
worlds & instances (of 10) & \descsetup{} & \exsetup{} & floor & \descsetup{} & \exsetup{} & floor \\
\midrule
0   & 0  & $34.6 \pm 6.4$ & $65.0 \pm 0.8$ & 52.1 & $98.3 \pm 0.9$ & $98.0 \pm 0.1$ & 75.4 \\
5   & 5  & $48.5 \pm 2.9$ & $70.0 \pm 0.6$ & 52.5 & $98.5 \pm 0.6$ & $98.1 \pm 0.3$ & 75.5 \\
25  & 10 & $48.0 \pm 7.4$ & $77.0 \pm 1.1$ & 53.2 & $98.4 \pm 0.1$ & $97.9 \pm 0.2$ & 74.1 \\
100 & 10 & $79.0 \pm 7.1$ & $86.6 \pm 1.6$ & 57.7 & $98.2 \pm 0.4$ & $97.8 \pm 0.3$ & 74.0 \\
228 & 10 & $79.0 \pm 5.4$ & $90.8 \pm 2.3$ & 56.7 & $97.8 \pm 0.9$ & $97.6 \pm 0.2$ & 68.8 \\
\bottomrule
\end{tabular}
\end{table}
 
\subsection{Further held-out sets for L3}
\label{app:results-splits}

Besides the L3 dial, we evaluate L3 on three further held-out sets, named by what their held-out instances have in common. The \emph{\splitmixed{}} split holds out rule instances spread across the pool, most of them supported. The \emph{\splitaxis{}} split holds out every column link, so that no trained rule combines the link type with the column axis, while row links remain in training. The \emph{\splitrowpair{}} split holds out the row links at one pair of row positions, so that no trained rule combines the link type with that position pair. For the \splitaxis{}, \splitrowpair{} and no-mix splits we make three independent draws of training and test worlds; the \splitmixed{} split has one draw. The no-mix split separates the value-shift rules (the link and shift families of Appendix~\ref{app:pool6}) from all other rules. Table~\ref{tab:splits} gives every draw. On every \splitaxis{} and \splitrowpair{} draw the \descsetup{} setup scores $9$ to $17$ points below the floor and the \exsetup{} setup $8$ to $15$ points above it, while both are tied with each other at L1 in the same runs.

A paired bootstrap over test worlds on draw $1$ (per-world accuracy averaged over all seeds of a setup, $10{,}000$ resamples) gives, for the \descsetup{} setup minus the floor, $-9.0$ with a $95\%$ interval of $[-13.0, -4.9]$ on the \splitaxis{} split and $-12.8$ $[-18.7, -6.8]$ on the \splitrowpair{} split; on no-mix L3 the same difference is $-2.9$ $[-10.6, +4.4]$, so its interval includes zero.

\begin{table}[htb]
\centering
\caption{Further held-out sets for L3, three draws each. Each cell gives L3 / L1 exact match in percent (for the no-mix split, L3 / L2). Mean $\pm$ standard deviation over $4$ seeds. The \splitmixed{} split has one draw: \descsetup{} $73.4 \pm 2.1$, \exsetup{} $81.2 \pm 0.9$, floor $55.9 \pm 0.3$ on L3.}
\label{tab:splits}
\footnotesize
\setlength{\tabcolsep}{4pt}
\begin{tabular}{@{}l c r r r@{}}
\toprule
Split & Draw & \descsetup{} & \exsetup{} & floor \\
\midrule
\splitaxis{}   & 1 & $36.4 \pm 3.7$ / 94.0 & $59.4 \pm 1.8$ / 97.0 & 45.4 / 66.3 \\
               & 2 & $37.3 \pm 3.2$ / 98.6 & $60.3 \pm 1.4$ / 97.8 & 50.4 / 69.5 \\
               & 3 & $35.8 \pm 2.1$ / 98.0 & $64.0 \pm 0.2$ / 97.8 & 50.8 / 76.8 \\
\addlinespace
\splitrowpair{} & 1 & $44.9 \pm 3.5$ / 97.3 & $72.2 \pm 0.5$ / 97.0 & 57.7 / 67.4 \\
               & 2 & $41.4 \pm 4.8$ / 97.8 & $69.6 \pm 1.3$ / 96.4 & 55.9 / 68.6 \\
               & 3 & $44.9 \pm 0.4$ / 98.8 & $69.4 \pm 1.6$ / 97.5 & 61.6 / 66.8 \\
\addlinespace
no-mix (L3 / L2) & 1 & $58.7 \pm 3.2$ / $67.6 \pm 5.5$ & $77.0 \pm 2.2$ / $82.3 \pm 1.6$ & 61.7 / 63.1 \\
               & 2 & $59.7 \pm 1.7$ / $67.6 \pm 2.1$ & $74.5 \pm 2.6$ / $78.1 \pm 4.5$ & 57.0 / 56.5 \\
               & 3 & $57.6 \pm 0.8$ / $66.4 \pm 2.5$ & $74.2 \pm 2.4$ / $79.5 \pm 3.7$ & 58.0 / 59.4 \\
\bottomrule
\end{tabular}
\end{table}
 
\subsection{Description dropout}
\label{app:dropout}

To test whether the \descsetup{} setup's deficit on unsupported instances comes from never having to infer rules from the grid, we train a variant in which the description prefix is replaced by null tokens on a random half of the training samples ($p = 0.5$). Such a model is evaluated both with the description present and with it masked; masked, it should behave like a \noinfo{} model of the same sequence length. Tables~\ref{tab:dropout} and~\ref{tab:masked} report both, on single splits. Three facts follow. First, an ordinary \descsetup{} model uses its description: masked, it falls to $28$ to $33\%$ on L3 at every level of the L3 dial and to about $40\%$ on L1 (Table~\ref{tab:masked}; masked L0 is $27$ to $36\%$ on every split and level). Second, dropout lifts the \descsetup{} model above the floor on every setting tried, at a cost of $1$ to $2$ points on L1. Third, on the L3 dial at level $0$, the dropout model evaluated with the description masked reproduces the $52$-token \noinfo{} model to within $2$ points on L3, L1 and L0, so the dropout model is a \noinfo{} model plus a description that only adds.

\begin{table}[htb]
\centering
\caption{Description dropout, evaluated with the description present. Exact match in percent, mean $\pm$ standard deviation over $4$ seeds. The floor column gives the \noinfo{} model at $36$ tokens and, after the slash, at $52$ tokens (padded to the \descsetup{} sequence length).}
\label{tab:dropout}
\footnotesize
\setlength{\tabcolsep}{4pt}
\begin{tabular}{@{}l r r r r@{}}
\toprule
Setting & \descsetup{} & + dropout & floor (36 / 52 tokens) & \exsetup{} \\
\midrule
L3 dial, level 0 (unsupported) & $34.6 \pm 6.4$ & $58.8 \pm 1.6$ & $52.1 \pm 2.3$ / $53.2 \pm 0.4$ & $65.0 \pm 0.8$ \\
\splitaxis{} draw 1 (unsupported) & $36.4 \pm 3.7$ & $53.9 \pm 2.2$ & $45.4 \pm 0.1$ / $46.0 \pm 2.0$ & $59.4 \pm 1.8$ \\
no-mix draw 1, L2 & $67.6 \pm 5.5$ & $71.8 \pm 2.2$ & $63.1 \pm 1.8$ / $64.0 \pm 1.3$ & $82.3 \pm 1.6$ \\
no-mix draw 1, L3 & $58.7 \pm 3.2$ & $66.6 \pm 2.1$ & $61.7 \pm 1.6$ / $62.9 \pm 0.2$ & $77.0 \pm 2.2$ \\
L3 dial, level 0, L1 & $98.3 \pm 0.9$ & $96.5 \pm 0.2$ & $75.4 \pm 0.9$ / $77.7 \pm 0.5$ & $98.0 \pm 0.1$ \\
\bottomrule
\end{tabular}
\end{table}
 \begin{table}[htb]
\centering
\caption{Evaluation with the description replaced by null tokens at test time, exact match in percent. Mean $\pm$ standard deviation over $4$ seeds. On the L3 dial at level $0$, the dropout model matches the $52$-token \noinfo{} model to within $2$ points; the ordinary \descsetup{} model collapses.}
\label{tab:masked}
\footnotesize
\setlength{\tabcolsep}{4pt}
\begin{tabular}{@{}l l r r r@{}}
\toprule
Setting & Level & \descsetup{}, masked & + dropout, masked & floor at 52 tokens \\
\midrule
L3 dial, level 0 & L3 & $28.0 \pm 2.9$ & $52.4 \pm 2.0$ & 53.2 \\
                 & L1 & $37.4 \pm 4.8$ & $78.4 \pm 0.6$ & 77.7 \\
                 & L0 & $33.3 \pm 3.8$ & $74.7 \pm 0.5$ & 72.8 \\
\bottomrule
\end{tabular}
\end{table}
 
\subsection{Controls}
\label{app:controls}

Table~\ref{tab:controls} gives the sequence-length and mechanism controls and Table~\ref{tab:mechanisms} the mechanism screen. Padding the \descsetup{} prefix to the \exsetup{} length ($108$ tokens, the same model size) makes the \splitaxis{} deficit worse and removes most of the \splitrowpair{} deficit, so sequence length is ruled out as the cause on the first split but not on the second. Further controls, draw $1$ of the \splitaxis{} split unless noted:
\begin{itemize}[itemsep=1pt,topsep=2pt,parsep=1pt,partopsep=0pt,leftmargin=*]
  \item \textbf{Halting.} Running all $32$ recurrent rounds instead of halting, over $50$ checkpoints and their L1 and L3 cells, changes the cell means by at most $1.4$ points in any setup and never raises them by more than $0.5$.
  \item \textbf{Training length.} Per-epoch L3 accuracy of the \descsetup{} setup stays between $28$ and $40\%$ from the first epoch on, with no upward trend; training accuracy passes $95\%$ at epoch $4$ and $99\%$ at epoch $9$, and L1 climbs to $94.6\%$. Training for $40$ epochs gives $34.4 \pm 2.8\%$ on L3, $94.5\%$ on L1 and $99.7\%$ on L0.
  \item \textbf{Training data.} With $1{,}000$ training worlds instead of $500$ and the same $10{,}000$ batches: \descsetup{} $43.1 \pm 3.1\%$ (L1 $98.0\%$), floor $47.2 \pm 0.4\%$ (L1 $67.7\%$).
  \item \textbf{Injection mechanism.} Five ways of injecting the description were screened on the \splitmixed{} split (Table~\ref{tab:mechanisms}); the three that train are within $2$ points on L1, and the best of them, composed rule vectors, gives $36.0 \pm 5.8\%$ on the \splitaxis{} split (Table~\ref{tab:controls}). In the design used in the main text the description is a prefix of the input: four tokens per rule, so $16$ description tokens placed before the $36$ puzzle cells and processed like any other input. The alternatives leave that prefix empty (null tokens) and hand the description to the model separately; the two that train are described below, and two further ones (a spatial field of learned grid patterns and FiLM-style per-block modulation) did not converge under the screening recipe (Table~\ref{tab:mechanisms}). All five run at $108$ positions under the screening recipe, and every added component starts as the identity.
  \begin{itemize}[itemsep=1pt,topsep=2pt,parsep=1pt,partopsep=0pt,leftmargin=*]
    \item \emph{Composed rule vectors.} A small network first merges the four tokens of each rule into one vector, so the description becomes four vectors, one per rule, added to the input at the first four positions. This tests whether the difficulty is keeping track of which tokens belong to the same rule.
    \item \emph{Cross-attention.} After each block, every one of the $36$ cells attends to the description tokens and reads from them, choosing by content which rule information to use, instead of relying on fixed position-to-position mixing weights.
  \end{itemize}
\end{itemize}

\begin{table}[htb]
\centering
\caption{Length and mechanism controls, draw $1$ of each split, L3 exact match in percent, $4$ seeds. Parameter counts: $1.09$M at $36$ tokens, $1.11$M at $52$, $1.25$M at $108$. Padding the \descsetup{} prefix to the \exsetup{} length moves the three hard splits by $-8.5$, $+10.4$ and $-0.2$ points, with no consistent sign. $^{a}$Screening recipe, $2$ seeds; the unpadded reference at that recipe is $67.9$. $^{b}$L2 on the same split: $58.1 \pm 8.2$. $^{c}$Screening recipe, $2$ seeds.}
\label{tab:controls}
\footnotesize
\setlength{\tabcolsep}{4pt}
\begin{tabular}{@{}l r r r r@{}}
\toprule
Control & \splitaxis{} & \splitrowpair{} & no-mix & \splitmixed{} \\
\midrule
\descsetup{} at 52 tokens (reference) & 36.4 & 44.9 & 58.7 & 73.4 \\
\descsetup{} padded to 108 tokens & $27.9 \pm 1.6$ & $55.3 \pm 5.0$ & $58.5 \pm 1.4$ & $67.5 \pm 3.8^{a}$ \\
\noinfo{} at 36 tokens & 45.4 & 57.7 & 61.7 & 55.9 \\
\noinfo{} at 52 tokens & $46.0 \pm 2.0$ & $58.8 \pm 0.0$ & $62.9 \pm 0.2$ & not run \\
composed rule vectors & $36.0 \pm 5.8$ & $57.9 \pm 10.2$ & $53.7 \pm 5.1^{b}$ & $70.8^{c}$ \\
\bottomrule
\end{tabular}
\end{table}
 \begin{table}[htb]
\centering
\caption{Injection mechanisms for the rule description, \splitmixed{} split, screening recipe ($12$ epochs, cosine decay), $2$ seeds. Exact match in percent at L0 / L1 / L3. The three working mechanisms are within $2$ points of each other on L1.}
\label{tab:mechanisms}
\footnotesize
\setlength{\tabcolsep}{4pt}
\begin{tabular}{@{}l r r r@{}}
\toprule
Mechanism & L0 & L1 & L3 \\
\midrule
composed rule vectors & 99.1 & 98.3 & 70.8 \\
cross-attention from cells to description tokens & 98.5 & 97.2 & 65.6 \\
description tokens in the sequence (padded) & 99.0 & 96.2 & 67.5 \\
spatial field (did not converge) & 92.4 & 91.5 & 60.5 \\
generated per-layer modulation (FiLM; failed) & 69.9 & 73.7 & 57.7 \\
\bottomrule
\end{tabular}
\end{table}
 
\subsection{Provenance of the earlier L1 result}
\label{app:provenance}

An earlier version of this study reported $66\%$ for the \descsetup{} setup at L1. Table~\ref{tab:provenance} crosses the old and new training code with the old and new split: the number follows the split and not the code. Independently, the ranking of setups depends on the recipe: under the screening recipe the \exsetup{} setup reaches only $75.7\%$ on L1 against $94.5\%$ for the \descsetup{} setup, with $20$ epochs and cosine decay $85.6$ against $98.5\%$, and under the paper recipe $95.4$ against $98.5\%$. An undertrained recipe can therefore reverse the comparison, and the earlier symbolic failure was a property of that split.

\begin{table}[htb]
\centering
\caption{Where the earlier L1 result for the \descsetup{} setup came from: training code (old / new) by split (old / new), L1 exact match in percent, $2$ seeds per cell except the top-left one ($25$ runs).}
\label{tab:provenance}
\footnotesize
\setlength{\tabcolsep}{4pt}
\begin{tabular}{@{}l r r@{}}
\toprule
 & old split & new split \\
\midrule
old training code & 65.8 & 95.0 \\
new training code & 66.5 & 98.5 \\
\bottomrule
\end{tabular}
\end{table}
 
\subsection{L3 accuracy by rule family}
\label{app:families}

Table~\ref{tab:families} breaks L3 down by the family of the held-out instance on the L2-dial split. At the natural exposure level the \descsetup{} setup beats the \exsetup{} setup on value-shift and uniqueness rules and loses only on parity rules, which alone account for the aggregate gap. Per held-out world at that level, the ten worlds whose held-out instance has every factor pair supported by at least one trained rule score $87$ to $100\%$ under both setups; the twenty worlds with an unsupported pair score $21$ to $94\%$ (\descsetup{}) against $54$ to $100\%$ (\exsetup{}), the parity ones lowest ($21$ to $69\%$ against $62$ to $80\%$).

\begin{table}[htb]
\centering
\caption{L3 exact match in percent by rule family of the held-out instance, L2-dial split, $4$ seeds, ten held-out worlds per family, at the lowest and the natural exposure level.}
\label{tab:families}
\footnotesize
\setlength{\tabcolsep}{4pt}
\begin{tabular}{@{}l l r r r r@{}}
\toprule
Exposure & Setup & value shift (links) & uniqueness & parity & aggregate \\
\midrule
$0\%$    & \descsetup{} & 75.8 & 66.3 & 34.2 & 58.7 \\
$0\%$    & \exsetup{}   & 93.2 & 74.6 & 63.1 & 77.0 \\
$74.6\%$ & \descsetup{} & 98.3 & 83.8 & 51.2 & 77.8 \\
$74.6\%$ & \exsetup{}   & 96.8 & 79.6 & 69.6 & 82.0 \\
\bottomrule
\end{tabular}
\end{table}
 
\subsection{Further limitations of the evidence}
\label{app:limitations}

\begin{itemize}[itemsep=1pt,topsep=2pt,parsep=1pt,partopsep=0pt,leftmargin=*]
  \item \textbf{One training recipe.} All setups share one recipe, and the ranking of setups is recipe-sensitive: under a shorter schedule the \exsetup{} setup is far from converged while the \descsetup{} setup is not (Appendix~\ref{app:provenance}). The L0 and L1 comparison therefore holds under the training budget used; a budget sweep is left to future work. L0 is also measured on the held-out puzzles of the worlds used for checkpoint selection, so it is a mildly optimistic in-distribution reference.
  \item \textbf{Single-split dials and controls.} Both exposure dials, the description-dropout results, and most controls are run on one draw of one split. The below-floor result itself is replicated on six draws of two splits, but the shape of the dials and the size of the dropout repair are not. On the L2 dial, about $8$ of the $31$ points the \descsetup{} setup gains are accounted for by the floor rising with exposure.
  \item \textbf{Behavioural evidence only.} The account of the description as a crutch rests on accuracies, masked evaluations and the dropout repair; we did not analyse what the model represents, so it is consistent with the evidence rather than established by it.
  \item \textbf{One description alphabet.} The symbolic description uses one factored scheme (type, axis, position, parameter). Three ways of injecting it agree to within $2$ points on L1 (Appendix~\ref{app:controls}), so the injection mechanism is not the bottleneck, but no other alphabet or factorization was tried.
\end{itemize}
 
\section{Extended discussion}
\label{app:discussion}

This appendix expands on how our parameterized-family framing relates to prior formal definitions of intelligence (extending Section~\ref{sec:related-work}).

\paragraph{Formal definitions of intelligence.} Legg and Hutter~\citep{legg2007universal} and Hern\'{a}ndez-Orallo~\citep{hernandez2017measure} define intelligence as aggregate performance over a distribution of environments or tasks, weighted toward simpler ones, but give no concrete, computable task family with an explicit notion of difficulty. Our parameterized family of rule combinations is such a family: difficulty is set by how many rules, and which ones, are held out, and generalization within the family becomes a measurable quantity; a recent proposal similarly argues for measuring skill acquisition without human cognition as the reference point~\citep{goldfeder2026aiembracespecializationsuperhuman}. In Chollet's terms~\citep{chollet2019arc}, levels L0 to L2 sit within local generalization (new points from a known distribution), while L3 is a small step toward the broad generalization his spectrum places well beyond current systems.
 
\section{Architecture and training specification}
\label{app:model}

This appendix completes Section~\ref{sec:model}: the full specification of the model and the training recipe. Appendix~\ref{app:design-choices} gives the rationale behind task and data design choices the main text states without justification.

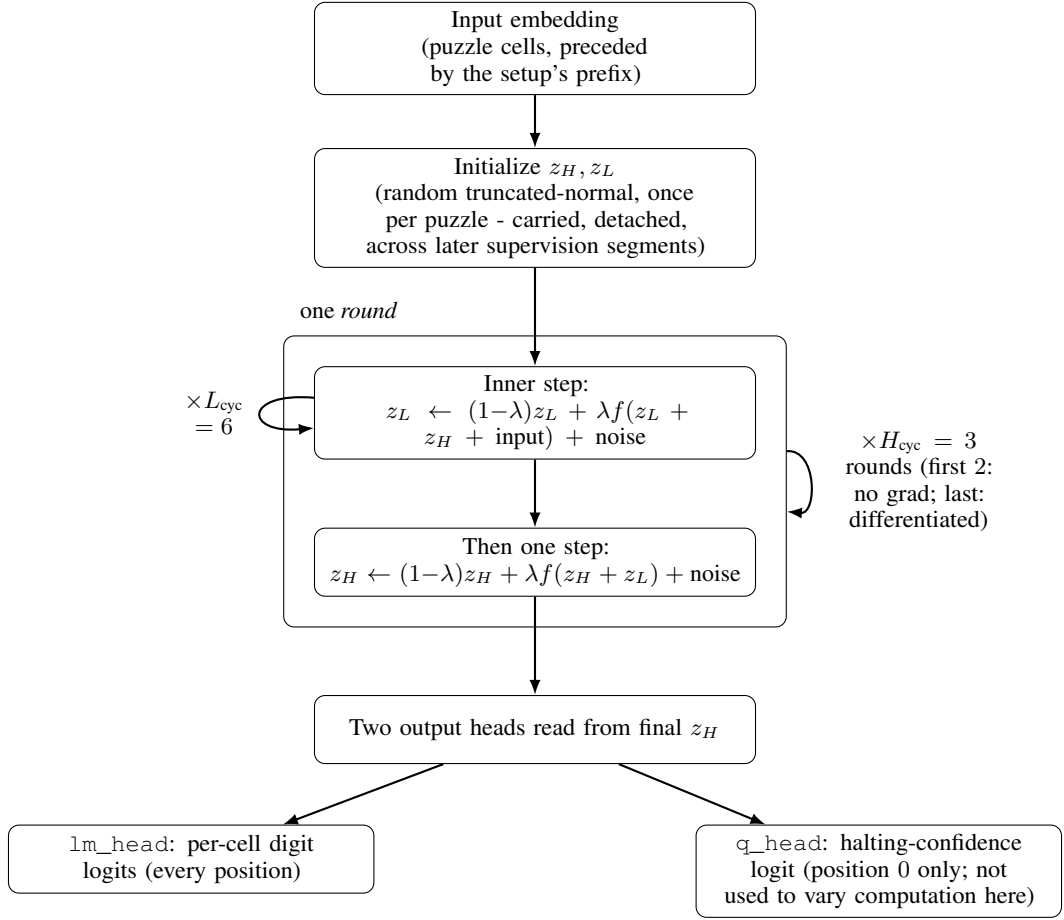
\begin{figure}[h]
\centering
\begin{tikzpicture}[
  box/.style={draw, rounded corners, align=center, minimum height=0.9cm, text width=5.6cm, font=\footnotesize, fill=white},
  arr/.style={-{Latex[length=2mm]}, thick},
  node distance=7mm
]
  \node[box] (input) {Input embedding\\(puzzle cells, preceded by the setup's prefix)};
  \node[box, below=of input] (init) {Initialize $z_H,z_L$\\(random truncated-normal, once per puzzle - carried, detached, across later supervision segments)};
  \node[box, below=13mm of init] (innerstep) {Inner step:\\$z_L \leftarrow (1{-}\lambda)z_L + \lambda f(z_L + z_H + \text{input}) + \text{noise}$};
  \node[box, below=9mm of innerstep] (zhupdate) {Then one step:\\$z_H \leftarrow (1{-}\lambda)z_H + \lambda f(z_H + z_L) + \text{noise}$};
  \node[box, below=13mm of zhupdate] (heads) {Two output heads read from final $z_H$};
  \node[box, below left=8mm and -8mm of heads, text width=4.6cm] (lm) {\texttt{lm\_head}: per-cell digit logits (every position)};
  \node[box, below right=8mm and -8mm of heads, text width=4.6cm] (q) {\texttt{q\_head}: halting-confidence logit (position $0$ only; not used to vary computation here)};
  \draw[arr] (input) -- (init);
  \draw[arr] (init) -- (innerstep);
  \draw[arr] (innerstep) -- (zhupdate);
  \draw[arr] (zhupdate) -- (heads);
  \draw[arr] (heads) -- (lm);
  \draw[arr] (heads) -- (q);
  \draw[arr] ([yshift=2mm]innerstep.west) to[out=175,in=185,looseness=6]
    node[left=1mm,align=center,font=\footnotesize] {$\times L_{\text{cyc}}$\\$=6$} ([yshift=-2mm]innerstep.west);
  \begin{scope}[on background layer]
    \node[draw, rounded corners, fit=(innerstep)(zhupdate), inner sep=4mm] (roundbox) {};
  \end{scope}
  \node[font=\footnotesize, anchor=south west] at ([xshift=1mm,yshift=1mm]roundbox.north west) {one \emph{round}};
  \draw[arr] ([yshift=4mm]roundbox.east) to[out=5,in=-5,looseness=1.3]
    node[right=1mm,align=center,font=\footnotesize,text width=2.4cm] {$\times H_{\text{cyc}}=3$ rounds (first $2$: no grad; last: differentiated)} ([yshift=-4mm]roundbox.east);
\end{tikzpicture}
\caption{The reasoning engine, used unchanged across the three setups; only the input sequence (top box) differs between them. $f$ is the $2$-layer, $256$-dimensional MLP-Mixer-style block stack described next, shared between the two update equations.}
\label{fig:backbone}
\end{figure}

Figure~\ref{fig:backbone} summarizes the overall flow described below.

\paragraph{Input.}
Tokens are embedded with a single embedding table shared by all setups, covering the $36$
symbols of Section~\ref{sec:how-rules-communicated}. The input is the conditioning prefix
followed by the $36$ cells of the puzzle: $16+36=52$ tokens for the \descsetup{} setup,
$2\cdot 36+36=108$ for the \exsetup{} setup, and $36$ for the \noinfo{} baseline.
Because the mixing weights are tied to absolute positions, no positional encoding is needed.

\paragraph{Recurrent core.}
Let $x$ be the embedded input and $(z_H, z_L)$ two latent states of the same shape,
initialised randomly. One pass performs $6$ updates of the fast state, each conditioned on
the slow state and the input, followed by one update of the slow state:
$z_L \leftarrow f(z_L, z_H + x)$, then $z_H \leftarrow f(z_H, z_L)$, where $f$ applies the
stack of $2$ blocks to the sum of its arguments. Updates are damped and noisy,
$h \leftarrow (1-\lambda)h + \lambda\,\mathrm{blocks}(h + \mathrm{inj}) + \epsilon$ with
$\lambda = 0.95$ and $\epsilon \sim \mathcal{N}(0, 0.01^2)$. Each forward call runs $3$ such
passes, of which only the last is differentiated; the earlier ones run without gradients.

\paragraph{Deep supervision and halting.}
Each training batch is processed in $8$ consecutive segments. After every segment the loss is
applied and an optimizer step is taken, and the latent states are carried to the next segment
detached from the graph, so gradients never span segments. Two heads read the slow state: a
linear head produces per-position logits over the vocabulary, of which only the last $36$
positions are scored, and a scalar halting head reads the first position. The loss is a
stablemax cross-entropy on the cells plus $0.5$ times a binary cross-entropy that trains the
halting head to predict whether the current prediction is exactly correct over all $36$ cells.
At evaluation the model is run for up to $32$ rounds; a puzzle halts the first time the halting
probability exceeds $0.9$, and the prediction at that round is frozen and scored.

\paragraph{Size.}
The model has hidden size $256$, $2$ blocks, and expansion factor $4$. The channel mixing,
the embedding and the heads are the same everywhere; only the token-mixing MLPs depend on the
sequence length, quadratically.

\paragraph{Optimization.}
AdamW, learning rate $10^{-4}$ with $\beta = (0.9, 0.95)$ and weight decay $0.01$, $200$
warmup steps followed by a constant rate, batch size $512$, $500$ batches per epoch and $20$
epochs, which is $10{,}000$ batches, each with $8$ optimizer steps (one per supervision segment), about five passes over the training puzzles. We
keep the checkpoint with the best validation exact-match accuracy. The same recipe is used for
every setup and every split; no per-setup tuning is performed. A shorter, decaying schedule leaves the \exsetup{} setup far from convergence while the \descsetup{} setup has converged, and reverses their ranking (Appendix~\ref{app:provenance}).

\paragraph{Deviations from HRM.}
Relative to \citet{wang2025hrm} we keep the hierarchical two-timescale recurrence, one-step-gradient training, and deep supervision, with three deviations: (i) attention blocks are replaced with MLP-Mixer-style~\citep{tolstikhin2021mlpmixer} token/channel-mixing blocks, since the grid topology and the number of prefix slots are fixed and known in advance, so a fixed mixing pattern suffices; (ii) a single supervised halting head replaces HRM's Q-learning ACT mechanism, and the update counts above stay fixed during training, with halting used only at evaluation; (iii) plain AdamW replaces HRM's specialized optimizer.
 
\section{Design choices}
\label{app:design-choices}

This appendix gives the rationale behind task and data design choices the main text (Section~\ref{sec:experimental-setup}) states without justification.

\paragraph{Why we chose $6\times6$ grids.}
Several pool rules tile the grid in $2\times3$ or $3\times2$ blocks, so the side length must be divisible by both $2$ and $3$; $6$ is the smallest such size. Compared with the classic $9\times9$, it keeps a sweep over thousands of rule combinations tractable without making puzzles trivial: the \noinfo{} baseline reaches only $61\%$ in-distribution exact-match accuracy on our main split, and at most $80\%$ on any split we use. Nothing in the formalization of Section~\ref{sec:worlds-framework} is specific to this size.

\paragraph{Why the base is fixed.}
Holding the base fixed isolates the object of study: any generalization gap reflects the variable pool, not an entangled second source of novelty. The base also makes puzzle generation tractable. Its grid-wide coupling (every row and column is an all-different group) is what lets the solver of Appendix~\ref{app:generation} propagate constraints and find uniquely completable partial configurations quickly, whereas most variable-pool rules are local (a box, a diagonal, one pair of rows) and constrain far fewer cells.

\paragraph{Why these rule types.}
Rules were chosen for their similarity to the base and to variants already familiar from Sudoku (boxes, diagonals, windows), each still introducing some difference, so the pool stays a single recognizable family rather than an arbitrary set. The pool is deliberately built so that most rules decompose into shared, reusable components (type, axis, position, parameter); the $30$ link rules in particular cover every row or column pair at every offset $k$, so that each component appears in several contexts (Appendix~\ref{app:pool6}).

\paragraph{Padding control.}
\label{app:padding}
Because the sequence length changes the size of the token-mixing MLPs and the compute per forward pass (Table~\ref{tab:setups}), we additionally run the \descsetup{} setup with its prefix padded by null tokens to $108$ positions, the \exsetup{} length, so that the two share the same token-mixing size and per-step compute, and the \noinfo{} setup padded to $52$ positions, the \descsetup{} length. Table~\ref{tab:controls} reports both: padding the description worsens its L3 deficit on the \splitaxis{} split and removes most of it on the \splitrowpair{} split, so sequence length is not a consistent explanation.

\end{document}